\documentclass[letterpaper]{article} % DO NOT CHANGE THIS
\usepackage{aaai2026}  % DO NOT CHANGE THIS
\usepackage{times}  % DO NOT CHANGE THIS
\usepackage{helvet}  % DO NOT CHANGE THIS
\usepackage{courier}  % DO NOT CHANGE THIS
\usepackage[hyphens]{url}  % DO NOT CHANGE THIS
\usepackage{graphicx} % DO NOT CHANGE THIS
\usepackage{natbib}  % DO NOT CHANGE THIS AND DO NOT ADD ANY OPTIONS TO IT
\usepackage{caption} % DO NOT CHANGE THIS AND DO NOT ADD ANY OPTIONS TO IT
\usepackage{algorithm}
\usepackage{algorithmic}
\usepackage{adjustbox}
\usepackage{booktabs}
\usepackage{multirow}
\usepackage{xcolor}
\usepackage{colortbl}
\usepackage{amssymb}
\usepackage{arydshln}
\usepackage{newfloat}
\usepackage{listings}
\DeclareCaptionStyle{ruled}{labelfont=normalfont,labelsep=colon,strut=off} % DO NOT CHANGE THIS
\floatstyle{ruled}
\newfloat{listing}{tb}{lst}{}
\floatname{listing}{Listing}
\title{Can Tool-Augmented Large Language Models \\Be Aware of Incomplete Conditions?}

\author{
Seungbin Yang\thanks{\hspace{0.2cm} Equal contribution}$^1$$^2$ \hspace{0.3cm} 
ChaeHun Park\footnotemark[1]$^1$ \hspace{0.3cm} 
Taehee Kim$^1$$^2$ \hspace{0.3cm} 
Jaegul Choo$^1$ 
 }

\affiliations{
$^1$ KAIST AI \hspace{0.3cm} $^2$ Letsur\\
\hspace{0cm}\texttt{\{sby99,ddehun,jchoo\}@kaist.ac.kr} \\
\hspace{0cm}\texttt{taehee.kim@letsur.ai} \\
}

\usepackage{bibentry}
\begin{document}

\maketitle

\begin{abstract}
Recent advancements in integrating large language models (LLMs) with tools have allowed the models to interact with real-world environments.
However, these tool-augmented LLMs often encounter incomplete scenarios when users provide partial information or the necessary tools are unavailable. Recognizing and managing such scenarios is crucial for LLMs to ensure their reliability, but this exploration remains understudied.
This study examines whether LLMs can identify incomplete conditions and appropriately determine when to refrain from using tools. 
To quantitatively evaluate this capability, we construct a new benchmark dataset where instances are systematically altered to simulate the ambiguous and incomplete conditions common in real-world interactions. 
Our experiments reveal that even state-of-the-art LLMs often struggle to identify these conditions, attempting to use tools without sufficient information or when the correct tool is unavailable.
To better understand these limitations, we conduct a detailed behavioral analysis across various conditions, including implicit evaluation and scenarios where models receive feedback from previous tool invocations.
Based on this analysis, we propose a novel prompting-based reasoning strategy that explicitly instructs models to assess the sufficiency of information and the availability of tools.
Our proposed approach significantly enhances the models' ability to recognize incomplete conditions, resulting in more informed and contextually appropriate tool-use decisions.
We believe our research contributes to advancing the reliability of LLMs, especially in real-world applications where incomplete or ambiguous information is common. Our dataset is available at \url{https://huggingface.co/datasets/ddehun/ICT}.
\end{abstract}

\section{Introduction}

Recently, there has been significant improvement in integrating large language models~(LLMs) with tools~\citep{li-etal-2023-api,qin2023toolllm,patil2023gorilla,schick2024toolformer,hao2024toolkengpt,liao-etal-2025-reflectool}.
These \textit{tool-augmented LLMs} can perceive up-to-date information, acquire real-world interaction capabilities, and perform complex tasks~\citep{tool-anyway}, enhancing user experiences across various applications~\citep{autogpt, hong2023metagpt}.

\begin{figure}[t!]
\centering
\includegraphics[width=0.9\columnwidth, height=6.2cm]{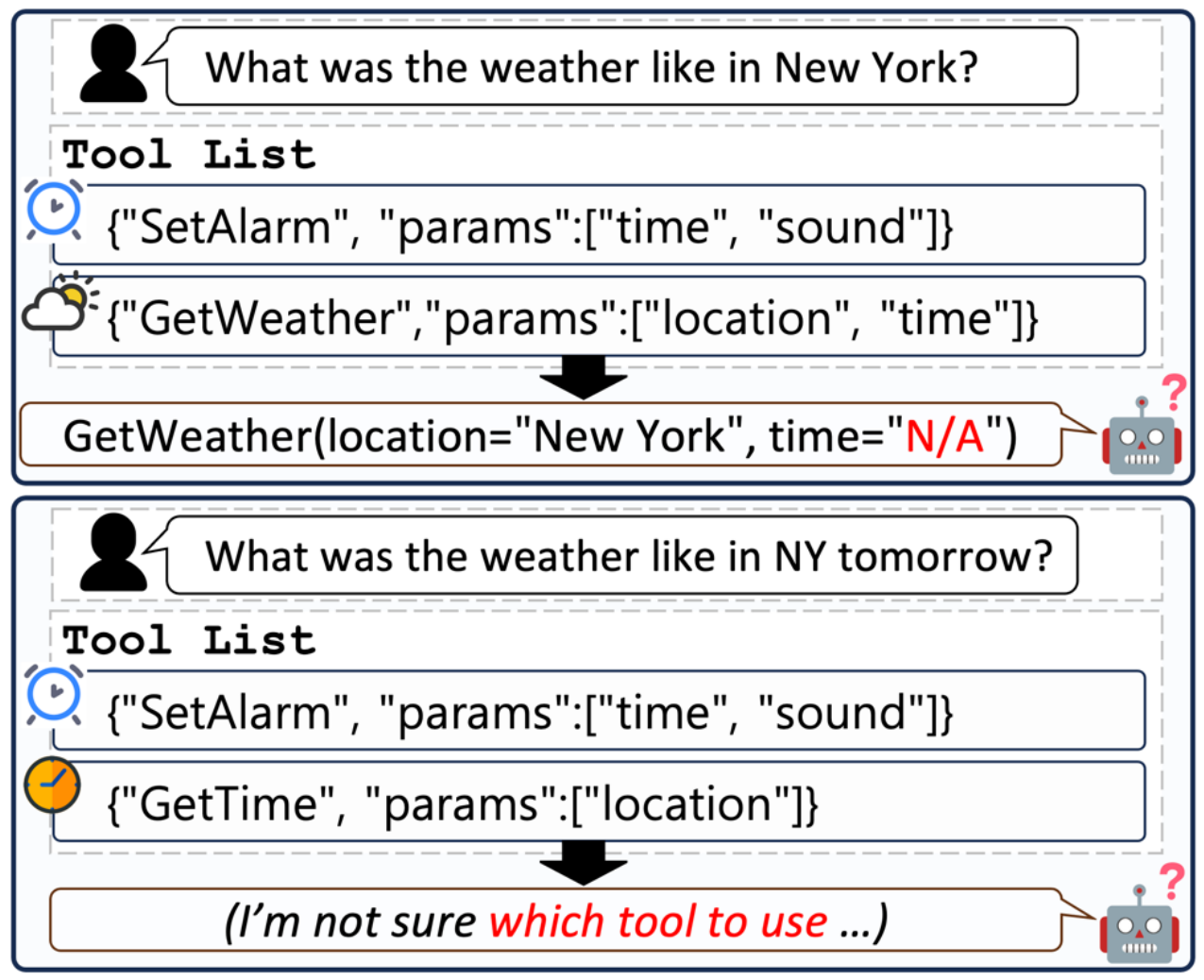}
% \caption{An illustration of incomplete conditions for tool invocation by LLMs.}
\caption{Illustration of incomplete conditions encountered by tool-augmented LLMs when invoking tools.}
\label{fig:mofit}
\end{figure}

Despite these advancements, tool-augmented LLMs often operate in situations where users lack sufficient knowledge of available tools or where the necessary tools do not exist.
As a result, LLM agents frequently encounter incomplete scenarios in which either a critical tool is missing or essential information for using it is not provided.
While previous work has contributed substantially to the development of tool-augmented LLMs, most existing approaches assume that all relevant tools and information are readily accessible~\citep{huang2023metatool, zhang-etal-2024-toolbehonest}.
We instead focus on whether tool-augmented LLMs can recognize when conditions are incomplete—either due to the absence of an appropriate tool or because the provided user input lacks sufficient detail for effective tool use—as illustrated in Figure~\ref{fig:mofit}.

To investigate this, we construct a benchmark by systematically modifying instances from existing tool-augmented LLM datasets~\citep{li-etal-2023-api,qin2023toolllm} and verifying each instance through human verification to ensure validity and naturalness. 
While related works such as BFCL V3~\citep{patil2025bfcl} and When2Call~\citep{ross-etal-2025-when2call} explore similar challenges, they either narrowly focus on specific tool domains or rely on a limited number of tools, thereby limiting their assessment of generalization across diverse real-world scenarios.

Our experiments reveal that state-of-the-art LLMs frequently struggle to recognize incomplete conditions, often allow tool invocations even when necessary information or tools are unavailable.
To explore these limitations, we conduct comprehensive analyses, including implicit evaluations and scenarios with feedback from prior tool interactions.
Motivated by models' frequent failure to accurately identify the necessary information for tool invocation, we introduce Structured Verification, a prompting-based approach that explicitly guides LLMs to systematically assess the adequacy of available information and the presence of required tools.
Our findings show that Structured Verification enhances the models’ ability to recognize incomplete conditions, leading to more accurate and contextually appropriate tool use.

Our contributions can be summarized as follows.
(1) We construct a dataset simulating incomplete conditions by manipulating tool-use datasets.
(2) We evaluate LLMs' ability to recognize impossible tool invocations, highlighting difficulties in identifying missing information.
(3) We comprehensively analyze model behavior in incomplete scenarios and demonstrate the effectiveness of Structured Verification.

\section{Related Work}
Recent research has explored the capability enhancement of LLMs with external tools~\citep{tool-anyway, gu-etal-2024-middleware, cai2024largelanguagemodelstool, fan-etal-2024-biasalert, liao-etal-2025-reflectool}, ranging from basic retrieval systems \citep{chen-etal-2017-reading} and arithmetic operations \citep{inaba-etal-2023-multitool, schick2024toolformer} to complex programming languages \citep{gou2024toratoolintegratedreasoningagent, zhang-etal-2024-codeagent} and APIs \citep{xu2023tool, guo2024stabletoolbench, yuan2024easytool}. Various benchmarks have been developed to evaluate LLMs' tool usage capabilities, including tool selection timing \citep{huang2023metatool}, robustness to noisy tool descriptions \citep{ye2024rotbench}, tool error handling \citep{sun-etal-2024-tools}, and safety considerations \citep{ye-etal-2024-toolsword}.

Although several studies have examined the awareness of LLMs when the necessary tools are not provided, the existing work has significant limitations.
% Previous studies focused on limited tool scenarios \citep{ning2024wtuevalwhetherornottoolusage, berkeley-function-calling-leaderboard}, assumed a perfect tool utilization plan is given \citep{huang-etal-2024-planning-creation}, or overlooked situations with similar but incorrect tools \citep{zhang-etal-2024-toolbehonest}. 
% \citet{lu2025toolsandbox} address only limited tool diversity and insufficient sample sizes, while \citet{ross-etal-2025-when2call} introduce a valuable benchmark for assessing LLMs in tool-use scenarios but do not incorporate rigorous human verification of their synthetically generated examples.
Previous studies focused on limited tool scenarios \citep{ning2024wtuevalwhetherornottoolusage, berkeley-function-calling-leaderboard, lu2025toolsandbox}, assumed a perfect tool utilization plan is given \citep{huang-etal-2024-planning-creation}, or overlooked situations with similar but incorrect tools \citep{zhang-etal-2024-toolbehonest}. 
\citet{ross-etal-2025-when2call} introduce a valuable benchmark for assessing LLMs in tool-use scenarios but do not incorporate rigorous human verification of their synthetically generated examples.
% Our work fills these gaps by exploring more realistic scenarios, including situations where incorrect but similar tools are provided or where users provide incomplete information required for proper tool calling.
Our work fills these gaps by exploring more diverse and realistic scenarios, validated through meticulous human annotation, including situations where incorrect but similar tools are provided or where users provide incomplete information required for proper tool calling.
Although research in question answering has explored handling irrelevant knowledge and ambiguous requests~\citep{min-etal-2020-ambigqa,kamath-etal-2020-selective,cole-etal-2023-selectively,adaptive-rag}, the ability of LLMs to recognize impractical tool usage still remains underexplored.

\section{How to Evaluate the Awareness of Tool-augmented LLMs}

\begin{figure}[t!]
    \centering
    \includegraphics[width=0.45\textwidth]{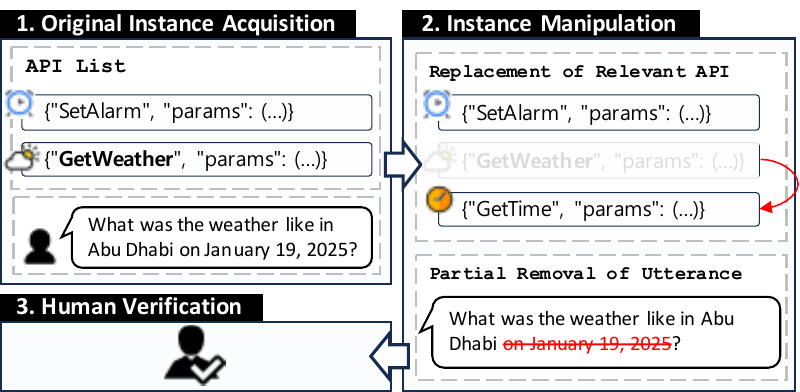}
    \caption{Dataset construction pipeline.}
    \label{fig:pipeline}
\end{figure}

To simulate incomplete scenarios where necessary tools are unavailable or users provide partial information, we manipulate instances from a test set of two benchmarks: APIBank \citep{li-etal-2023-api} and ToolBench \citep{qin2023toolllm}.
Both datasets are designed to evaluate how effectively LLMs can respond to user requests using APIs.\footnote{We refer to Tool and API interchangeably.}
In their original dataset, each instance has an available API that can address the user's request, with sufficient information provided to invoke APIs.
We deliberately manipulate these instances to simulate scenarios with missing tools or incomplete user-supplied information.
% This section describes our data source, manipulation strategy, and human verification process for creating reliable incomplete scenarios. 
This section describes our data source, manipulation strategy, and human verification process. 
Fig. \ref{fig:pipeline} illustrates the overall pipeline of our dataset construction.

\begin{figure*}
\centering
\includegraphics[width=0.95\textwidth]{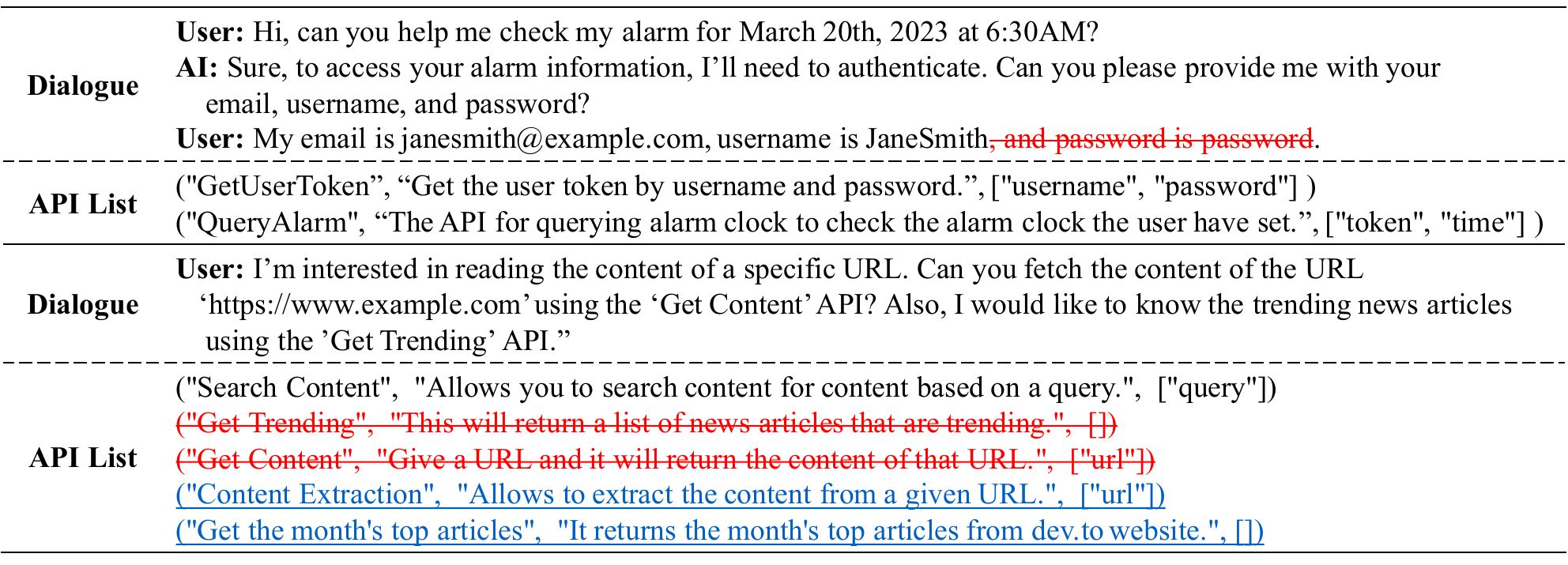}
\caption{
\textbf{Data Examples}. The upper example illustrates a case of utterance removal in APIBank, while the lower example shows an API replacement from the ToolBench dataset. Removed and newly introduced information, as part of our manipulation strategy, are highlighted with a strikethrough and an underline, respectively. 
% Each API is represented as a tuple containing API name, description, and parameters. 
}
\label{tab:main_sample_tables}
\end{figure*}

\subsection{Data Source}
\label{secsec:data_source}

We construct a new dataset by leveraging two existing datasets: APIBank \citep{li-etal-2023-api} and ToolBench \citep{qin2023toolllm}.
APIBank consists of 73 APIs and 314 annotated multi-turn conversations. We select 450 instances from the test split, excluding those that require a tool-retrieval module. ToolBench is based on 16,000 real-world APIs across 49 categories from RapidAPI Hub.\footnote{\url{https://rapidapi.com/hub}} We utilize 764 test instances as filtered by \citet{guo2024stabletoolbench}. These instances are used as the original data for manipulation.

\subsection{Simulating Incomplete Scenarios}
\label{secsec:data_manipulation}
We simulate two incomplete scenarios in which LLMs cannot properly invoke tools by manipulating the original dataset instances: (1) replacing relevant APIs with irrelevant ones, and (2) partially removing user utterances.

\paragraph{Replacement of Relevant API} 
We use a dense retriever to replace the appropriate APIs with similar but irrelevant ones.
This manipulation ensures that the desired tool is unavailable, simulating a scenario where LLMs must decide not to use any of the provided tools.
Specifically, the relevant APIs in the original instance are replaced with other APIs.
To find semantically similar APIs, we use the sentence encoder proposed by \citet{gao-etal-2021-simcse}.\footnote{\url{princeton-nlp/sup-simcse-roberta-large}} We concatenate the name and description of each API, convert this text into a fixed-size vector, and then select one of the most similar APIs by calculating cosine similarity between the relevant API and all available APIs in the dataset. 
This strategy reflects realistic situations where LLMs encounter APIs that appear semantically similar but are unsuitable due to subtle differences in functionality or required parameters.

\paragraph{Partial Removal of User Utterance}
We partially remove user utterances in a conversation to mimic scenarios where (1) the request is unclear, making it impossible to use APIs, or (2) the request is clear but lacks necessary information for API invocation. This manipulation removes essential information from user utterances, rendering appropriate API invocation infeasible.
We automate this process using a proprietary LLM (i.e., GPT-4 \citep{achiam2023gpt}) to generate naturally corrupted dialogues.
The model is instructed to identify and remove crucial information required for tool invocation, supported by reasoning instructions and five manually-crafted few-shot examples.
% The model is instructed to identify and remove crucial information required for tool invocation, and we enhance the quality of this process with reasoning prompts and five manually crafted few-shot examples.
% Sample instances are in the supplementary material. % in Fig.~\ref{fig:app:apibank_utterance_removal_success} to Fig.~\ref{fig:app:toolbench_utterance_removal_wrong}.

\begin{table}[t!]
    \centering
    \setlength{\tabcolsep}{1mm}
    % \begin{adjustbox}{width=0.45\textwidth}
    \begin{tabular}{lcccc}
        \toprule
         & \multicolumn{2}{c}{\textbf{API Replacement}} & \multicolumn{2}{c}{\textbf{Utterance Removal}} \\     
         & APIBank & ToolBench & APIBank & ToolBench \\ 
        \midrule
        \# Instances & 423 & 477 & 304 & 406 \\ \midrule
        Avg. Turns & 6.18 & 1 & 5.18 & 1 \\ 
        Avg. APIs & 2.13 & 5.13 & 2.05 & 5.51 \\ 
        Avg. Uttr.
        % Avg. Uttr. Word
        & 17.66 & 52.38 & 18.87 & 51.15 \\
        Avg. API
        % Avg. API Length
        & 434.39 & 713.35 & 443.38 & 747.53 \\
        \bottomrule
    \end{tabular}
    % \end{adjustbox}
    \caption{\textbf{Dataset Statistics}. The \textit{API Replacement} and \textit{Utterance Removal} denote \textit{Relevant API Replacement} and \textit{Partial Removal of User Utterance}, respectively. 
    \textit{Avg. Uttr.} denotes the average word count per utterance and \textit{Avg. API} denotes the average string length of each API.}
    % \textit{Avg. API Length} denotes the string length of each API.}
    \label{app:tab:stat_table}
\end{table}
\subsection{Data Verification}
\label{secsec:data_verification}

To ensure dataset validity, we manually reviewed all instances to remove cases where user requests could still be fulfilled by the provided APIs or where the manipulations resulted in unnatural.
Two authors holding bachelor's degrees or higher in Computer Science conducted this verification process, focusing on: (1) whether alternative or non-replaced APIs could still handle the user's request in API Replacement cases, and (2) whether the remaining information was sufficient for API execution in Utterance Removal cases.\footnote{Detailed verification criteria are in the supplementary material.}
% \ref{sec:app:verification_guideline}.} 
Through active discussion and refinement of filtering criteria, we ensured that the final dataset includes only instances that genuinely represent realistic incomplete scenarios.

\subsection{Dataset Statistics}
Our final dataset consists of 727 instances from APIBank and 883 from ToolBench, totaling 1,610 instances. Among these, 900 instances were generated by replacing relevant APIs, while 710 were created by removing parts of the user utterances.
% Table \ref{app:tab:stat_table} presents the overall dataset statistics, and Fig. \ref{tab:main_sample_tables} provides examples of different dataset manipulation strategies.
Table \ref{app:tab:stat_table} presents the overall dataset statistics, and Fig. \ref{tab:main_sample_tables} provides examples of dataset manipulation strategies.

% \usepackage{multirow}
\begin{table*}[t]
\centering
\setlength{\tabcolsep}{1mm} 
\begin{adjustbox}{width=\textwidth}
\begin{tabular}{ccccccccccccccc}
\toprule
 &  &  & 
 % \multicolumn{1}{c}{\begin{tabular}[c]{@{}c@{}}Mistral\\ 7B\end{tabular}} & 
 \multicolumn{1}{c}{\begin{tabular}[c]{@{}c@{}}Llama3.1\\ 8B\end{tabular}} & \multicolumn{1}{c}{\begin{tabular}[c]{@{}c@{}}Llama3.1\\ 70B\end{tabular}} & 
 % \multicolumn{1}{c}{\begin{tabular}[c]{@{}c@{}}Claude3\\ haiku\end{tabular}} & 
 \multicolumn{1}{c}{\begin{tabular}[c]{@{}c@{}}Gemma\\ 3-4b\end{tabular}} &
 \multicolumn{1}{c}{\begin{tabular}[c]{@{}c@{}}Gemma\\ 3-27b\end{tabular}} &
 \multicolumn{1}{c}{\begin{tabular}[c]{@{}c@{}}Qwen2.5\\ 7B\end{tabular}} &
 \multicolumn{1}{c}{\begin{tabular}[c]{@{}c@{}}Qwen2.5\\ 14B\end{tabular}} &
 \multicolumn{1}{c}{\begin{tabular}[c]{@{}c@{}}Qwen2.5\\ 32B\end{tabular}} &
 \multicolumn{1}{c}{\begin{tabular}[c]{@{}c@{}}Qwen2.5\\ 72B\end{tabular}} &
 \multicolumn{1}{c}{\begin{tabular}[c]{@{}c@{}}GPT-3.5\\ Turbo\end{tabular}} & \multicolumn{1}{c}{GPT-4} & \multicolumn{1}{c}{\begin{tabular}[c]{@{}c@{}}GPT-4o\\ mini\end{tabular}} & \multicolumn{1}{c}{GPT-4o} \\
 \hline
  \multicolumn{15}{l}{\textit{\textbf{Relevant API Replacement}}} \\
 \hline
\multirow{4}{*}{APIBank} & \multirow{2}{*}{Acc.} & 0-shot & 
  76.06 & 79.10 &   67.85 & 78.49 & 73.05 & 80.97 & 85.82 & \underline{87.94}&
67.92 & 87.85 & \textbf{88.82} & 86.03 \\
 &  & 4-shot &  75.70 & 82.50 &  69.50 & 76.12 & 78.01 & 80.50 & 84.75 & 87.35& 71.32 & 85.66 & 85.66 & 83.35 \\  
 & \multirow{2}{*}{F1} & 0-shot   & 71.32 & 81.22 &  57.10 & 73.31 & 65.97 & 77.35 & 83.83 & 87.02&62.39 & \underline{87.56} & \textbf{88.41} & 86.23 \\
 &  & 4-shot   & 74.81 & 83.33 &   58.25 & 69.76 & 72.81 & 76.99 & 82.73 & 86.54&65.80 & 84.99 & 85.99 & 82.90 \\ \midrule
 
\multirow{4}{*}{ToolBench} & \multirow{2}{*}{Acc.} & 0-shot  & 69.45 & 72.64 &   55.56 & 64.36 & 72.33 & 69.29 & 70.55 & 74.11& 64.29 & 72.31 & 77.69 & 76.48 \\ 
 &  & 4-shot   & 76.37 & \textbf{80.55}  & 57.65 & 72.64 & 72.12 & 74.53 & 72.43 & 73.48&57.91 & \underline{79.01} & 75.82 & 77.14 \\ 
 & \multirow{2}{*}{F1} & 0-shot  & 56.43 & 76.97 &20.90 & 45.51 & 64.80 & 57.96 & 58.74 & 67.80&46.28 & 61.93 & 72.97 & 75.06 \\
 &  & 4-shot   & 70.34 & \textbf{80.49} & 28.62 & 64.39 & 64.15 & 67.98 & 75.53 & 75.51&25.34 & \underline{77.02} & 69.36 & 76.89 \\
 \hline
 \multicolumn{15}{l}{\textit{\textbf{Partial Removal of User Utterance}}} \\ \hline
\multirow{4}{*}{APIBank} & \multirow{2}{*}{Acc.} & 0-shot  & 52.63 & 77.42 &  53.12 & 62.66 & 54.77 & 79.28 & 81.09 & 81.25&56.20 & \underline{89.47} & 77.25 & 88.29 \\
 &  & 4-shot & 59.25 & 84.72 & 51.48 & 77.96 & 56.74 & 80.92 & 79.77 & 86.02& 65.87 & 88.79 & 79.46 & \textbf{89.81} \\ 
 & \multirow{2}{*}{F1} & 0-shot   & 19.60 & 80.06  & 21.49 & 42.24 & 26.67 & 74.80 & 77.76 & 77.29&36.76 & \underline{89.27} & 73.41 & 88.74 \\ 
 &  & 4-shot & 46.90 & 85.98  & 12.98 & 72.87 & 27.55 & 77.78 & 75.74 & 84.79&55.43 & 88.70 & 79.10 &\textbf{ 90.07} \\ \midrule
\multirow{4}{*}{ToolBench} & \multirow{2}{*}{Acc.} & 0-shot & 50.19 & 71.54 & 50.00 & 52.09 & 50.86 & 50.99 & 54.68 & 61.45&49.42 & 57.05 & 57.57 & \underline{74.51} \\
 &  & 4-shot   & 51.36 & 72.83 &  50.99 & 52.46 & 50.12 & 51.11 & 50.62 & 64.41&50.32 & 66.11 & 55.11 & \textbf{75.16} \\
 & \multirow{2}{*}{F1} & 0-shot  & 5.41 & \textbf{74.94 }  &1.46 & 10.16 & 5.23 & 15.68 & 21.37 & 43.19& 2.98 & 27.19 & 35.18 & 70.90 \\
 &  & 4-shot & 14.93 & 70.25 & 7.44 & 15.35 & 2.88 & 13.88 & 11.87 & 51.91& 2.04 & 55.74 & 25.05 & \underline{73.33} \\
 \bottomrule
\end{tabular}
\end{adjustbox}
\caption{Performance evaluation results of LLM by manipulation type. The accuracy (Acc.) and F1 score (F1) are used for evaluation metrics. Both the zero-shot and four-shot performance are presented. The highest and the second-highest scores in each metric are highlighted in \textbf{bold} and \underline{underlined}.}
\label{tab:main}
\end{table*}

\section{Experiments}
\label{sec:experiments}

\subsection{Setup}
\label{secsec:setup}

% We note that our binary classification setup aligns well with multi-stage decision-making processes in real-world applications, where a model first decides whether tool invocation is feasible before proceeding \citep{huang2023metatool, huang-etal-2024-planning-creation}.

\paragraph{Task Formulation}
% We aim to evaluate whether LLMs can detect incomplete conditions, where tool invocation should be withheld due to missing information or unavailable tools.
% To this end, we formulate the task as a binary classification problem, asking the model to determine whether tool use is feasible based on the given dialogue and available APIs.
We investigate whether LLMs can recognize incomplete conditions, where tool invocation should be withheld due to insufficient user input or unavailable tools. 
We frame this as a binary classification task, asking the model to determine whether tool use is feasible based on the given dialogue and available APIs.
This setup provides a clear evaluation signal and aligns with real-world multi-stage decision-making, where models must first assess the viability of tool usage before proceeding \citep{huang2023metatool, huang-etal-2024-planning-creation}. We use both original and manipulated instances from APIBank and ToolBench, prompting the model to answer ``Yes” or ``No” accordingly, and evaluate performance using accuracy and F1 score, with “No” as the positive class.

\paragraph{Models}
% The following open-source and proprietary LLMs are used for our experiments:
In our experiments, we employ the following open-source and proprietary LLMs:
% Phi-3-small-8k-instruct \citep{abdin2024phi}, 
% Mistral-Instruct-v0.2 (7B)~\citep{jiang2023mistral}, 
Llama3.1-8B/70B-Instruct \citep{dubey2024llama}, 
Gemma3-4b/27b-it \citep{team2025gemma}
Qwen2.5-7B/14B/32B/72B-Instruct \citep{yang2024qwen2.5}, 
% Claude-3-Haiku~\citep{anthropic2024claude}, 
% GPT-3.5-Turbo~\citep{openai2023chatgpt}, and GPT-4~\citep{achiam2023gpt}, GPT-4o-mini \citep{achiam2023gpt}, and GPT-4o \citep{achiam2023gpt}.
GPT-3.5-Turbo~\citep{openai2023chatgpt}, GPT-4, GPT-4o-mini, and GPT-4o \citep{achiam2023gpt}.

\paragraph{Implementation Details}
\label{sec:app:impl_detail}
We implement all open-source LLMs with the Transformers library \citep{wolf-etal-2020-transformers}.
For proprietary models, we use 
% \texttt{claude-3-haiku-20240307}, 
\texttt{gpt-3.5-turbo-0125},  \texttt{gpt-4-0613}, \texttt{gpt-4o-mini-2024-07-18}, and \texttt{gpt-}\texttt{4o-2024-11-20} for 
% Claude-3-Haiku, 
GPT-3.5-Turbo, GPT-4, GPT-4o-mini, and GPT-4o, respectively. The temperature is set to 0 across all models.
We use two original and two manipulated instances to evaluate with few-shot examples.
In experiments with instances from ToolBench, we select models with a context length exceeding 8192 tokens to handle the frequently encountered lengthy API descriptions.

\subsection{Results}
Our experiments reveal several insights regarding the capabilities and limitations of tool-augmented LLMs in recognizing incomplete conditions as follows.

% \noindent\textbf{Model performance depends on API complexity and compositional conversational context.}
\noindent\textbf{Model performance depends on API complexity and conversational context.}
Models exhibit significant performance differences across manipulation types and datasets (Table \ref{tab:main}). GPT-4o-mini and Qwen2.5-72B achieve strong accuracy (88.82\% and 87.94\%, respectively) in the API Replacement scenario on APIBank. However, their performance notably declines in scenarios requiring compositional reasoning from conversational context, especially on ToolBench, reflecting their sensitivity to complex API structures.

\noindent\textbf{Models frequently fail to abstain from inappropriate tool invocations.} LLMs consistently struggle to abstain from invoking tools under incomplete conditions, as indicated by low F1 scores (e.g., as low as 5.41\% for Llama3.1-8B on ToolBench Utterance Removal). This tendency to overestimate tool applicability highlights a significant limitation in models' awareness of information sufficiency.

\noindent\textbf{Few-shot demonstrations have limited impact on complex incompleteness recognition.} Few-shot learning generally enhances performance on APIBank, yet has minimal impact on ToolBench scenarios. This suggests that simple demonstrations are insufficient for models to robustly identify incomplete conditions~\citep{khot2022decomposed}.

These findings underline critical challenges for LLMs in reliably detecting incomplete scenarios, especially those involving complex APIs and nuanced conversational contexts, highlighting the need for structured reasoning approaches.

% \noindent{\textbf{Model performance varies significantly across manipulation types and datasets.}} Experimental results in Table \ref{tab:main} show distinct performance patterns. In API Replacement, GPT-4 variants and Llama 3.1 70B achieve strong results (88.82\% and 79.10\% accuracy on APIBank). The performance gap becomes wider in the Utterance Removal condition, particularly for smaller models on ToolBench. Extremely low F1 scores in ToolBench are due to models predominantly predicting ``Yes," demonstrating their tendency to overestimate tool applicability. This suggests that handling Utterance Removal tasks effectively requires a deeper understanding of incomplete contexts.

% \noindent{\textbf{Few-shot learning shows dataset-dependent effectiveness.}} The impact of few-shot learning varies across datasets. While models generally perform better with few-shot examples on APIBank, improvements are less pronounced in ToolBench. We attribute this to the complexity of real-world tools, which require significant reasoning skills and involve diverse scenarios \citep{khot2022decomposed}.

\begin{table}[t!]
\centering
\setlength{\tabcolsep}{1mm} 
% \begin{adjustbox}{width=0.95\columnwidth}
\begin{tabular}{lcccccc}
\toprule
          & \multicolumn{2}{c}{\textbf{API}}     & \multicolumn{2}{c}{\textbf{Utterance}}  &\multicolumn{2}{c}{\textbf{Overall}} \\ 
&Acc. &F1 &Acc. &F1 &Acc. &F1  \\ \midrule
Human  & 81.11  &78.48&81.11 &81.32 &  81.11 & 80.00    \\ \bottomrule
\end{tabular}
% \end{adjustbox}
\caption{Human analysis results on our evaluation set.}
\label{tab:human_analysis}
\end{table}

\subsection{Analysis}

\paragraph{Human Analysis}
We conducted a human analysis on the same tasks to establish a performance baseline for LLM evaluation. Specifically, we selected 180 instances from the APIBank dataset, ensuring class balance. Nine participants with relevant academic backgrounds were recruited to solve ``Yes/No" questions for 20 instances each, following the same evaluation protocol applied to the LLMs. Each participant was compensated with \$10 for their participation.

Results in Table \ref{tab:human_analysis} show consistent human performance (81.11\% accuracy) across manipulation types, with F1 scores from 78.48\% to 81.32\%.
This provides a meaningful reference point for model evaluation, with some larger models demonstrating comparable performance.
% This provides a meaningful reference point for model evaluation, with some larger models demonstrating comparable or superior performance in handling incomplete situations.

\begin{table}[t]
\centering
\setlength{\tabcolsep}{1mm} 
% \begin{adjustbox}{width=\columnwidth}
\begin{tabular}{lccc}
\toprule
\textbf{Dataset} & \textbf{\# Domains} & \textbf{\# Tools} & \textbf{\# Examples} \\
\midrule
\textbf{BFCL v3} & 8 & 76 & 400 \\
\textbf{ToolSandbox} & 11 & 34 & 224 \\
\textbf{When2Call} & 5 & 504 & 2,357 \\
\textbf{Ours} & 57 (49) & 2,799 (2750) & 1,610 (883) \\
\bottomrule
\end{tabular}
% \end{adjustbox}

\caption{Comparison of datasets used to evaluate incomplete tool usage. Our dataset provides broader coverage across domains and tools. Numbers in parentheses indicate the respective contributions from ToolBench.}
\label{tab:dataset_comparison}
\end{table}

\begin{figure}[t!]
\centering
\includegraphics[width=\columnwidth]{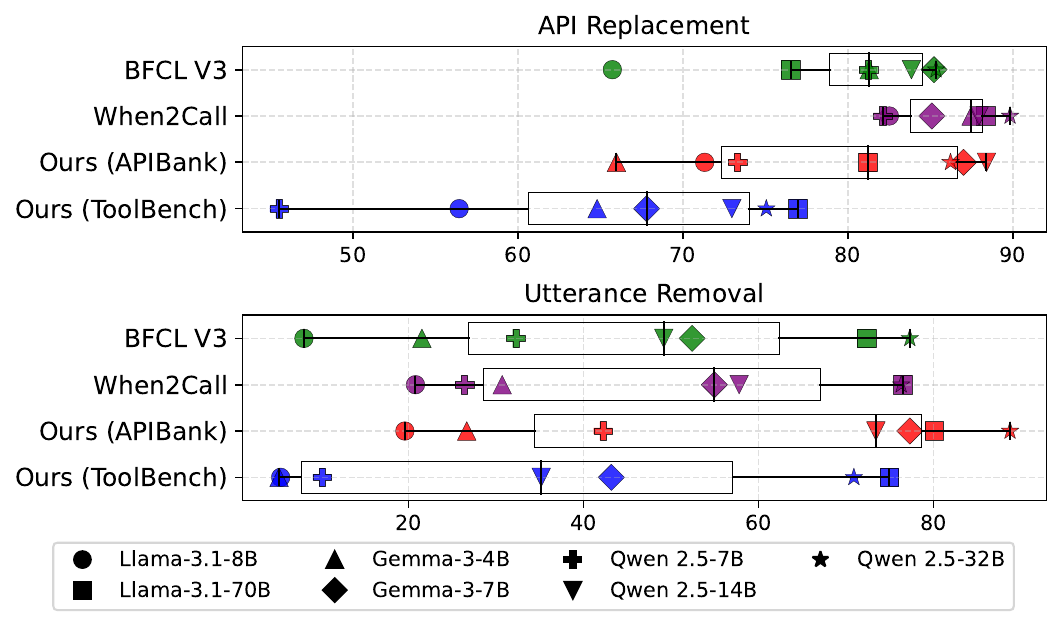}
\caption{F1-score comparison across different benchmarks for simulating incomplete scenarios for tool usage.}
\label{fig:data_comparison_pdf}
\end{figure}

\paragraph{Comparison with Prior Tool Use Datasets}
Several datasets have been proposed to evaluate LLMs under incomplete tool usage conditions, including BFCL v3 \citep{patil2025bfcl}, ToolSandBox \citep{lu2025toolsandbox}, and When2Call \citep{ross-etal-2025-when2call}.
As summarized in Table~\ref{tab:dataset_comparison}, our dataset provides broader coverage by incorporating and modifying instances from both APIBank and ToolBench, resulting in a larger set of domains (57) and tools (2,799). 
Moreover, in addition to its broader scope in domains and tools, our dataset is distinguished by its quality assurance; human annotators have manually verified every constructed instance to ensure its validity.

To compare performance across these benchmarks, we evaluate a suite of LLMs in a consistent zero-shot setting.
As shown in Fig.~\ref{fig:data_comparison_pdf}, the F1 scores for all models are generally lowest on our dataset. 
This performance drop is particularly sharp on the subset derived from ToolBench, especially in the Utterance Removal.
This suggests that the evaluation contexts from our benchmark, particularly the ToolBench subset, are more complex for models to resolve. 
While lower performance does not by itself indicate higher dataset quality, the consistent degradation across all models suggests our benchmark effectively captures a broad and challenging range of real-world failure modes for incomplete conditions.

\begin{table}[t!]
\centering
\setlength{\tabcolsep}{1mm} 
% \begin{adjustbox}{width=0.95\columnwidth}
\begin{tabular}{lcccccc}
\toprule
 & \multicolumn{3}{c}{\textbf{API Replacement}} & \multicolumn{3}{c}{\textbf{Utterance Removal}} \\ 
 & Acc. & F1 & Corr. & Acc. & F1 & Corr. \\ \midrule
Llama3.1-8B	&58.27&	39.66&	68.56&	60.36&	45.1&	76.48 \\
Llama3.1-70B	&75.13	&68.34	&81.23&	79.61&	75.0&	84.81 \\
% GPT-3.5-Turbo & 59.69 & 52.17 & 58.75 & 59.70 & 49.90 & 62.66 \\
GPT-4 & 80.50 & 77.18 & 85.22 & 87.83 & 86.75 & 88.32 \\
\bottomrule
\end{tabular}
% \end{adjustbox}
\caption{Implicit evaluation results on the APIBank dataset. The Correlation (Corr.) metric measures the instance-wise agreement between the model's binary classification and free-form generation predictions.}
\label{tab:freeform_table}
\end{table}

\paragraph{Implicit Evaluation with Free-form Generation}
\label{app:sec:implicit_experiments}
As described in the experiments section, our experimental setup involves a binary classification task in which models are asked to determine whether the current conditions are incomplete for tool invocation by answering ``Yes" or ``No." This approach directly assesses the models' ability to evaluate whether the necessary conditions for tool usage are met. 
We also consider an alternative method that implicitly gauges the models' awareness to complement our explicit evaluation.
In this setup, the models engage in a simulated dialogue with a user, and their next actions are monitored. If the model generates a response that includes invoking a tool, it is considered that the model has determined that it can invoke the tool.
In contrast, the absence of tool invocation indicates that the model has determined that it cannot invoke the tool. We apply this implicit evaluation framework to Llama3.1-8B/70B and GPT-4 on the APIBank dataset.

The results presented in Table \ref{tab:freeform_table} indicate that models generally perform worse in implicit evaluation setups. For example, GPT-4's F1 scores drop from 87.56\% and 89.27\% to 77.18\% and 86.75\% for API Replacement and Utterance Removal, respectively.
Based on these results, we hypothesize that the binary classification setup, which explicitly prompts the model to assess whether the current tool list or user information is sufficient for tool invocation, enables the model to better recognize when tool usage is appropriate.

\begin{table}[t]
\centering
\setlength{\tabcolsep}{1mm} 
\begin{adjustbox}{width=1\columnwidth}
\begin{tabular}{cc|ccccc}
\toprule
Utterance  & Errors in     & Llama   &Llama   & GPT     & GPT    & GPT   \\ 
Removal    &  API Call      & 3.1-8B & 3.1-70B  & 4   & 4o-mini &  4o \\ 
    
 \midrule
\checkmark                & \checkmark       &89.0	&83.0                & 86.0    & 97.0 & 96.0       \\
\checkmark                 &    	&40.0	&82.0& 67.0           & 45.0 & 79.0\\
                 & \checkmark          	&2.0&	92.0              & 99.0   & 51.0 & 91.0\\    \bottomrule   
\end{tabular}
\end{adjustbox}
\caption{The results of providing API invocation outcomes to LLMs across different scenarios. Accuracy is used as the evaluation metric, highlighting the differences in how different models handle erroneous API call results and their ability to recognize hallucinations and incomplete conditions.}
\label{tab:appendix_table_task3}
\end{table}

\begin{figure*}[t!]
\centering
\includegraphics[width=\textwidth]{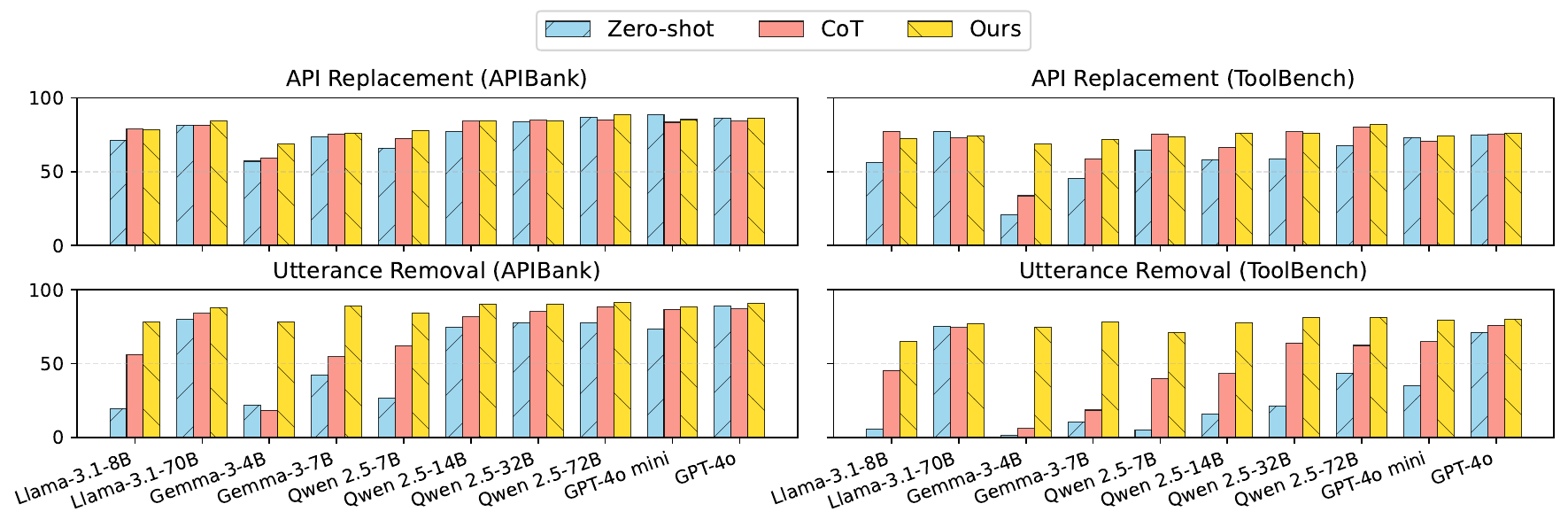}
\caption{F1-score comparison across different manipulation types and datasets, showing the impact of CoT and Structured Verification (Ours) prompting. Full results are in the supplementary material. }
% Table \ref{tab:cot_full}.}
\label{fig:cot_results}
\end{figure*}

\begin{figure}[t]
\centering
\includegraphics[width=\columnwidth]{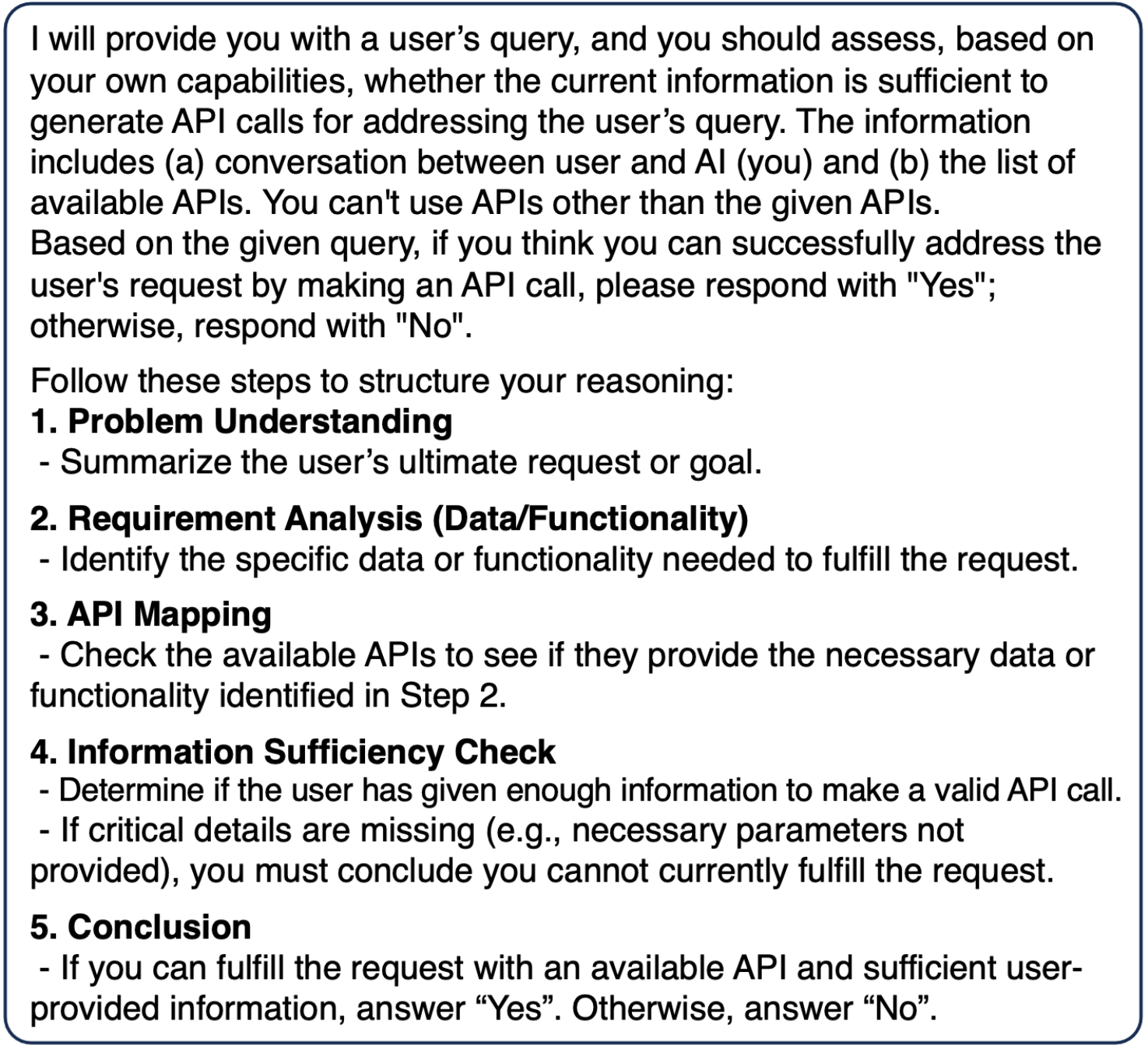}
\caption{Structured Verification Reasoning Process. 
% The full prompt is in the supplementary material.
}
% supplemenFigure \ref{fig:app:ours_prompt_template}.}

\label{fig:main:verification_prompt}
\end{figure}

\paragraph{Impact of Tool Invocation Feedback on Incomplete Condition Recognition}
\label{app:sec:self-verification}
Our primary experiments assess whether LLMs can recognize incomplete conditions prior to tool invocation, a practical approach for preventing incorrect or unsafe tool use.
However, in more permissive settings, observing the outcomes of tool invocations may help models reassess input completeness.
Prior work suggests that utilizing execution feedback can improve reasoning and planning abilities~\citep {react, qin2023toolllm, gou2024criticlargelanguagemodels}.

To investigate this, we sample 100 manipulated utterance removal instances and their original counterparts from APIBank. For each instance, we simulate two types of API outcomes: (1) \textit{Error}, where the API call fails due to missing information; and (2) \textit{Success}, where the call executes correctly. Based on this, we construct three evaluation scenarios: (i) \textit{Utterance Removal + Wrong API Call}, where the call fails due to insufficient user input; (ii) \textit{Utterance Removal + Hallucinated API Call}, where the call succeeds using hallucinated content not grounded in the dialogue; and (iii) \textit{Complete Condition + Wrong API Call}, where sufficient user input is present but the call still fails. In each case, the model is given the API list, dialogue history, and API call result, and is asked to judge whether the provided information is sufficient to fulfill the user request.

Results show that Llama3.1-8B tends to infer incompleteness whenever the call result is an error, regardless of input sufficiency (Table~\ref{tab:appendix_table_task3}). In contrast, GPT-4 and GPT-4o more accurately distinguish between truly incomplete inputs and other error types. Furthermore, in cases where hallucinated API calls succeed, Llama3.1-8B and GPT-4o-mini often fail to detect the discrepancy, whereas models including Llama3.1-70B and GPT-4o reliably identify the missing information. These findings suggest that stronger reasoning capabilities enable more accurate interpretation of tool outcomes and improved recognition of incomplete conditions.

\begin{figure*}[t!]
\centering
\includegraphics[width=\textwidth]{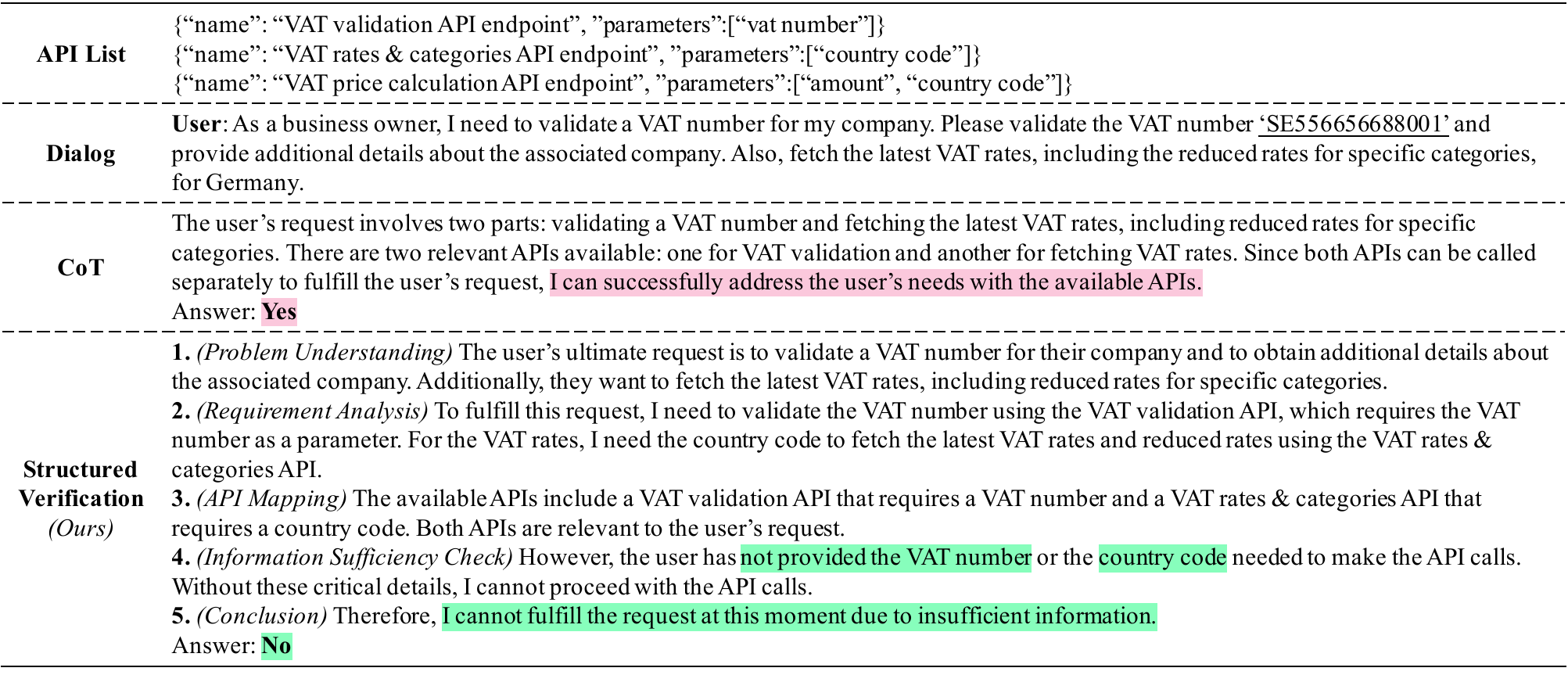}
\caption{\textbf{Case Study}. The example illustrates the corrupted sample generated by our utterance removal strategy in the ToolBench dataset. Removed information, as part of our manipulation strategy, is highlighted with \underline{underline}. Wrong and Correct reasoning from the models is manually highlighted. GPT-4o-mini is used as an evaluated LLM.
% is evaluated using both \textit{CoT} and \textit{Structured Verification} prompting strategies.
}
\label{fig:qualitative}
\end{figure*}

\section{Structured Verification for Handling Incomplete Conditions}
\subsection{Proposed Method}
% Our experiments reveal a critical weakness in how tool-augmented LLMs handle incomplete conditions.
% Models frequently overlook the need to assess whether the user input contains all necessary information, leading them to initiate tool use based on incomplete or underspecified conditions. 
% Binary classification 대신 direct하게 응답하는거나 Tool execution results를 제공하는 것도 완전하지는 않았다.
% This tendency to bypass preliminary validation results in unreliable API calls and suboptimal performance, as the models do not inherently structure their reasoning to assess the sufficiency of the given context.
% This deficit in pre-execution planning and analysis is a key reason for failure in scenarios with incomplete information.
Our experiments reveal a critical weakness in how tool-augmented LLMs handle incomplete conditions.
LLMs often overlook whether the user input contains sufficient information, leading them to invoke tools based on under-specified or partial contexts.
Neither prompting models to respond directly nor providing tool execution results fully resolves this issue, as both approaches often fail to enforce structured assessment of input sufficiency.
This tendency to skip preliminary validation results in unreliable tool use and degraded performance, as models lack an inherent reasoning process to evaluate contextual completeness.
This deficiency in pre-execution planning and reasoning is a key factor behind failures in incomplete scenarios.

Inspired by prior work highlighting the effectiveness of explicit self-assessment~\citep{manakul2023selfcheckgpt,shinn2023reflexion,dhuliawala2023chain} and structured decomposition for improving reasoning reliability~\citep{press2023measuring,ouyang2024structured}, we adapt these principles to tool-augmented settings, where systematically evaluating input completeness before tool invocation is essential.
Specifically, we propose Structured Verification, a method that enforces a systematic pre-execution check.
It compels the model first to decompose the user's request into its core requirements and then explicitly verify whether the available information and tools are sufficient to meet them.
Our prompting strategy guides the model through the following key steps: (1) identifying the user’s ultimate goal, (2) enumerating the required information or functionalities, (3) checking available tools for alignment with those requirements, and (4) verifying information sufficiency.
Fig. \ref{fig:main:verification_prompt} illustrates the system prompt designed for this process.
% Motivated by recent work on self-verification~\citep{manakul2023selfcheckgpt,shinn2023reflexion,dhuliawala2023chain} and structured reasoning in LLMs~\citep{press2023measuring,ouyang2024structured}, we adapt similar principles to tool-augmented settings, where verifying input sufficiency is critical for safe and accurate tool use.
% In response, we propose Structured Verification, a method that enforces a systematic pre-execution check. It compels the model first to decompose the user's request into its core requirements and then explicitly verify whether the available information and tools are sufficient to meet them. Our prompting strategy guides the model through the following key steps: (1) identifying the user’s ultimate goal, (2) enumerating the required information or functionalities, (3) checking available tools for alignment with those requirements, and (4) verifying information sufficiency. Fig. \ref{fig:main:verification_prompt} illustrates the system prompt designed for this process.

By explicitly prompting the model to decompose the problem and assess resource availability, our structured verification process improves decision-making in situations involving incomplete or partial information. Furthermore, this strategy operates without requiring tool execution or task-specific fine-tuning, making it a robust and efficient solution for high-stakes settings or scenarios where models must continuously adapt to newly introduced tools.

\subsection{Experiments}
\paragraph{Setup}
We evaluate our method using our manipulated dataset constructed from APIBank and ToolBench.
In this setup, we compare Chain-of-Thought (CoT) prompting with our Structured Verification method. CoT prompts the LLM to \textit{"Begin your response with brief reasoning first," } encouraging implicit reasoning without a formal structure. The full CoT prompt is shown in the supplementary material.
% Fig. \ref{fig:app:cot_prompt_template}.

\paragraph{Results}
As shown in Fig. \ref{fig:cot_results}, both CoT and Structured Verification significantly improve performance over zero-shot prompting across all models. In the API Replacement case, for instance, Gemma3-4B’s F1 score improves from 20.90\% to 33.68\% with CoT and further increases to 68.57\% with Structured Verification on ToolBench.
Similar trends are consistently observed across other models, where CoT boosts performance, particularly for smaller models, and Structured Verification shows further improvement in most cases. Overall, these results highlight that while prompting for reasoning is broadly beneficial, the explicit decomposition and checking enforced by Structured Verification provide a more robust path to the correct conclusion.

\paragraph{Case Study} Fig. \ref{fig:qualitative} presents a case study illustrating how different reasoning strategies affect the model's behavior when dealing with missing information. This example comes from the ToolBench dataset, where our utterance removal strategy introduces a corrupted version of the original user utterance.
The missing details, such as the VAT number and country code, are critical for making the API calls. GPT-4o-mini, evaluated with both CoT and Structured Verification prompting, exhibits different behaviors.
With CoT prompting, the model incorrectly assumes it can proceed without the missing information and mistakenly concludes that the request can be fulfilled, as highlighted by the red-colored box.
 However, with Structured Verification, the model correctly identifies the missing details and concludes that the request cannot be fulfilled, as shown in the green-colored box.
 This case study demonstrates how Structured Verification handles incomplete queries more effectively by explicitly checking for required information.

\section{Conclusion}
\label{sec:conclusion}
In this work, we investigate whether tool-augmented LLMs can recognize incomplete conditions before invoking tools. To enable systematic evaluation, we construct a benchmark by manipulating and human-verifying instances from APIBank and ToolBench. Our findings show that even strong models often fail to abstain from tool use when crucial information or tools are missing. To address this, we propose Structured Verification, a prompting strategy that guides models to explicitly assess input sufficiency and tool availability. This study offers not only a practical solution but also a comprehensive analysis framework for understanding and evaluating the robustness of tool-augmented LLMs under incomplete and ambiguous conditions.

% In this work, we investigate the ability of tool-augmented large language models (LLMs) to handle incomplete conditions, such as when users provide partial information or when required tools are inaccessible.
% Our experiments, based on carefully manipulated datasets, reveal varying performance patterns across different incomplete scenarios and model scales.
% Larger models show promising results, but the effectiveness of enhancement strategies like few-shot learning varies depending on the manipulation type and dataset.
% We also introduce Structured Verification, a method that explicitly guides models through a step-by-step evaluation process, which improves handling of incomplete information and leads to more reliable outcomes.

\bibliography{aaai2026}

\begin{thebibliography}{52}
\providecommand{\natexlab}[1]{#1}

\bibitem[{Achiam et~al.(2023)Achiam, Adler, Agarwal, Ahmad, Akkaya, Aleman, Almeida, Altenschmidt, Altman, Anadkat et~al.}]{achiam2023gpt}
Achiam, J.; Adler, S.; Agarwal, S.; Ahmad, L.; Akkaya, I.; Aleman, F.~L.; Almeida, D.; Altenschmidt, J.; Altman, S.; Anadkat, S.; et~al. 2023.
\newblock Gpt-4 technical report.
\newblock \emph{arXiv preprint arXiv:2303.08774}.

\bibitem[{Cai et~al.(2024)Cai, Wang, Ma, Chen, and Zhou}]{cai2024largelanguagemodelstool}
Cai, T.; Wang, X.; Ma, T.; Chen, X.; and Zhou, D. 2024.
\newblock Large Language Models as Tool Makers.
\newblock In \emph{The Twelfth International Conference on Learning Representations}.

\bibitem[{Chen et~al.(2017)Chen, Fisch, Weston, and Bordes}]{chen-etal-2017-reading}
Chen, D.; Fisch, A.; Weston, J.; and Bordes, A. 2017.
\newblock Reading {W}ikipedia to Answer Open-Domain Questions.
\newblock In Barzilay, R.; and Kan, M.-Y., eds., \emph{Proceedings of the 55th Annual Meeting of the Association for Computational Linguistics (Volume 1: Long Papers)}, 1870--1879. Vancouver, Canada: Association for Computational Linguistics.

\bibitem[{Cole et~al.(2023)Cole, Zhang, Gillick, Eisenschlos, Dhingra, and Eisenstein}]{cole-etal-2023-selectively}
Cole, J.; Zhang, M.; Gillick, D.; Eisenschlos, J.; Dhingra, B.; and Eisenstein, J. 2023.
\newblock Selectively Answering Ambiguous Questions.
\newblock In Bouamor, H.; Pino, J.; and Bali, K., eds., \emph{Proceedings of the 2023 Conference on Empirical Methods in Natural Language Processing}, 530--543. Singapore: Association for Computational Linguistics.

\bibitem[{Dhuliawala et~al.(2024)Dhuliawala, Komeili, Xu, Raileanu, Li, Celikyilmaz, and Weston}]{dhuliawala2023chain}
Dhuliawala, S.; Komeili, M.; Xu, J.; Raileanu, R.; Li, X.; Celikyilmaz, A.; and Weston, J. 2024.
\newblock Chain-of-Verification Reduces Hallucination in Large Language Models.
\newblock In \emph{Findings of the Association for Computational Linguistics ACL 2024}, 3563--3578.

\bibitem[{Dubey et~al.(2024)Dubey, Jauhri, Pandey, Kadian, Al-Dahle, Letman, Mathur, Schelten, Yang, Fan et~al.}]{dubey2024llama}
Dubey, A.; Jauhri, A.; Pandey, A.; Kadian, A.; Al-Dahle, A.; Letman, A.; Mathur, A.; Schelten, A.; Yang, A.; Fan, A.; et~al. 2024.
\newblock The llama 3 herd of models.
\newblock \emph{arXiv preprint arXiv:2407.21783}.

\bibitem[{Fan et~al.(2024)Fan, Chen, Xu, and Liu}]{fan-etal-2024-biasalert}
Fan, Z.; Chen, R.; Xu, R.; and Liu, Z. 2024.
\newblock {B}ias{A}lert: A Plug-and-play Tool for Social Bias Detection in {LLM}s.
\newblock In Al-Onaizan, Y.; Bansal, M.; and Chen, Y.-N., eds., \emph{Proceedings of the 2024 Conference on Empirical Methods in Natural Language Processing}, 14778--14790. Miami, Florida, USA: Association for Computational Linguistics.

\bibitem[{Gao, Yao, and Chen(2021)}]{gao-etal-2021-simcse}
Gao, T.; Yao, X.; and Chen, D. 2021.
\newblock {S}im{CSE}: Simple Contrastive Learning of Sentence Embeddings.
\newblock In Moens, M.-F.; Huang, X.; Specia, L.; and Yih, S. W.-t., eds., \emph{Proceedings of the 2021 Conference on Empirical Methods in Natural Language Processing}, 6894--6910. Online and Punta Cana, Dominican Republic: Association for Computational Linguistics.

\bibitem[{Gou et~al.(2024{\natexlab{a}})Gou, Shao, Gong, Yang, Duan, Chen et~al.}]{gou2024criticlargelanguagemodels}
Gou, Z.; Shao, Z.; Gong, Y.; Yang, Y.; Duan, N.; Chen, W.; et~al. 2024{\natexlab{a}}.
\newblock CRITIC: Large Language Models Can Self-Correct with Tool-Interactive Critiquing.
\newblock In \emph{The Twelfth International Conference on Learning Representations}.

\bibitem[{Gou et~al.(2024{\natexlab{b}})Gou, Shao, Gong, Yang, Huang, Duan, Chen et~al.}]{gou2024toratoolintegratedreasoningagent}
Gou, Z.; Shao, Z.; Gong, Y.; Yang, Y.; Huang, M.; Duan, N.; Chen, W.; et~al. 2024{\natexlab{b}}.
\newblock ToRA: A Tool-Integrated Reasoning Agent for Mathematical Problem Solving.
\newblock In \emph{The Twelfth International Conference on Learning Representations}.

\bibitem[{Gu et~al.(2024)Gu, Shu, Yu, Liu, Dong, Tang, Srinivasa, Latapie, and Su}]{gu-etal-2024-middleware}
Gu, Y.; Shu, Y.; Yu, H.; Liu, X.; Dong, Y.; Tang, J.; Srinivasa, J.; Latapie, H.; and Su, Y. 2024.
\newblock Middleware for {LLM}s: Tools Are Instrumental for Language Agents in Complex Environments.
\newblock In Al-Onaizan, Y.; Bansal, M.; and Chen, Y.-N., eds., \emph{Proceedings of the 2024 Conference on Empirical Methods in Natural Language Processing}, 7646--7663. Miami, Florida, USA: Association for Computational Linguistics.

\bibitem[{Guo et~al.(2024)Guo, Cheng, Wang, Liang, Qin, Li, Liu, Sun, and Liu}]{guo2024stabletoolbench}
Guo, Z.; Cheng, S.; Wang, H.; Liang, S.; Qin, Y.; Li, P.; Liu, Z.; Sun, M.; and Liu, Y. 2024.
\newblock StableToolBench: Towards Stable Large-Scale Benchmarking on Tool Learning of Large Language Models.
\newblock In \emph{Findings of the Association for Computational Linguistics ACL 2024}, 11143--11156.

\bibitem[{Hao et~al.(2024)Hao, Liu, Wang, and Hu}]{hao2024toolkengpt}
Hao, S.; Liu, T.; Wang, Z.; and Hu, Z. 2024.
\newblock Toolkengpt: Augmenting frozen language models with massive tools via tool embeddings.
\newblock \emph{Advances in neural information processing systems}, 36.

\bibitem[{Hong et~al.(2024)Hong, Zhuge, Chen, Zheng, Cheng, Wang, Zhang, Wang, Yau, Lin et~al.}]{hong2023metagpt}
Hong, S.; Zhuge, M.; Chen, J.; Zheng, X.; Cheng, Y.; Wang, J.; Zhang, C.; Wang, Z.; Yau, S. K.~S.; Lin, Z.; et~al. 2024.
\newblock MetaGPT: Meta Programming for A Multi-Agent Collaborative Framework.
\newblock In \emph{The Twelfth International Conference on Learning Representations}.

\bibitem[{Huang et~al.(2024)Huang, Zhong, Lu, Zhu, Gao, Liu, Hou, Zeng, Wang, Shang, Jiang, Xu, and Liu}]{huang-etal-2024-planning-creation}
Huang, S.; Zhong, W.; Lu, J.; Zhu, Q.; Gao, J.; Liu, W.; Hou, Y.; Zeng, X.; Wang, Y.; Shang, L.; Jiang, X.; Xu, R.; and Liu, Q. 2024.
\newblock Planning, Creation, Usage: Benchmarking {LLM}s for Comprehensive Tool Utilization in Real-World Complex Scenarios.
\newblock In Ku, L.-W.; Martins, A.; and Srikumar, V., eds., \emph{Findings of the Association for Computational Linguistics ACL 2024}, 4363--4400. Bangkok, Thailand and virtual meeting: Association for Computational Linguistics.

\bibitem[{Huang et~al.(2023)Huang, Shi, Li, Fan, Wu, Zhang, Liu, Zhou, Wan, Gong et~al.}]{huang2023metatool}
Huang, Y.; Shi, J.; Li, Y.; Fan, C.; Wu, S.; Zhang, Q.; Liu, Y.; Zhou, P.; Wan, Y.; Gong, N.~Z.; et~al. 2023.
\newblock MetaTool Benchmark: Deciding Whether to Use Tools and Which to Use.
\newblock In \emph{The Twelfth International Conference on Learning Representations}.

\bibitem[{Inaba et~al.(2023)Inaba, Kiyomaru, Cheng, and Kurohashi}]{inaba-etal-2023-multitool}
Inaba, T.; Kiyomaru, H.; Cheng, F.; and Kurohashi, S. 2023.
\newblock {M}ulti{T}ool-{C}o{T}: {GPT}-3 Can Use Multiple External Tools with Chain of Thought Prompting.
\newblock In Rogers, A.; Boyd-Graber, J.; and Okazaki, N., eds., \emph{Proceedings of the 61st Annual Meeting of the Association for Computational Linguistics (Volume 2: Short Papers)}, 1522--1532. Toronto, Canada: Association for Computational Linguistics.

\bibitem[{Jeong et~al.(2024)Jeong, Baek, Cho, Hwang, and Park}]{adaptive-rag}
Jeong, S.; Baek, J.; Cho, S.; Hwang, S.~J.; and Park, J.~C. 2024.
\newblock Adaptive-RAG: Learning to Adapt Retrieval-Augmented Large Language Models through Question Complexity.
\newblock In \emph{Proceedings of the 2024 Conference of the North American Chapter of the Association for Computational Linguistics: Human Language Technologies (Volume 1: Long Papers)}, 7029--7043.

\bibitem[{Kamath, Jia, and Liang(2020)}]{kamath-etal-2020-selective}
Kamath, A.; Jia, R.; and Liang, P. 2020.
\newblock Selective Question Answering under Domain Shift.
\newblock In Jurafsky, D.; Chai, J.; Schluter, N.; and Tetreault, J., eds., \emph{Proceedings of the 58th Annual Meeting of the Association for Computational Linguistics}, 5684--5696. Online: Association for Computational Linguistics.

\bibitem[{Khot et~al.(2022)Khot, Trivedi, Finlayson, Fu, Richardson, Clark, and Sabharwal}]{khot2022decomposed}
Khot, T.; Trivedi, H.; Finlayson, M.; Fu, Y.; Richardson, K.; Clark, P.; and Sabharwal, A. 2022.
\newblock Decomposed Prompting: A Modular Approach for Solving Complex Tasks.
\newblock In \emph{The Eleventh International Conference on Learning Representations}.

\bibitem[{Li et~al.(2023)Li, Zhao, Yu, Song, Li, Yu, Li, Huang, and Li}]{li-etal-2023-api}
Li, M.; Zhao, Y.; Yu, B.; Song, F.; Li, H.; Yu, H.; Li, Z.; Huang, F.; and Li, Y. 2023.
\newblock {API}-Bank: A Comprehensive Benchmark for Tool-Augmented {LLM}s.
\newblock In Bouamor, H.; Pino, J.; and Bali, K., eds., \emph{Proceedings of the 2023 Conference on Empirical Methods in Natural Language Processing}, 3102--3116. Singapore: Association for Computational Linguistics.

\bibitem[{Liao et~al.(2025)Liao, Jiang, Wang, and Wang}]{liao-etal-2025-reflectool}
Liao, Y.; Jiang, S.; Wang, Y.; and Wang, Y. 2025.
\newblock {R}eflec{T}ool: Towards Reflection-Aware Tool-Augmented Clinical Agents.
\newblock In Che, W.; Nabende, J.; Shutova, E.; and Pilehvar, M.~T., eds., \emph{Proceedings of the 63rd Annual Meeting of the Association for Computational Linguistics (Volume 1: Long Papers)}, 13507--13531. Vienna, Austria: Association for Computational Linguistics.
\newblock ISBN 979-8-89176-251-0.

\bibitem[{Lu et~al.(2025)Lu, Holleis, Zhang, Aumayer, Nan, Bai, Ma, Ma, Li, Yin et~al.}]{lu2025toolsandbox}
Lu, J.; Holleis, T.; Zhang, Y.; Aumayer, B.; Nan, F.; Bai, H.; Ma, S.; Ma, S.; Li, M.; Yin, G.; et~al. 2025.
\newblock ToolSandbox: A Stateful, Conversational, Interactive Evaluation Benchmark for LLM Tool Use Capabilities.
\newblock In \emph{Findings of the Association for Computational Linguistics: NAACL 2025}, 1160--1183.

\bibitem[{Manakul, Liusie, and Gales(2023)}]{manakul2023selfcheckgpt}
Manakul, P.; Liusie, A.; and Gales, M. 2023.
\newblock SelfCheckGPT: Zero-Resource Black-Box Hallucination Detection for Generative Large Language Models.
\newblock In \emph{The 2023 Conference on Empirical Methods in Natural Language Processing}.

\bibitem[{Min et~al.(2020)Min, Michael, Hajishirzi, and Zettlemoyer}]{min-etal-2020-ambigqa}
Min, S.; Michael, J.; Hajishirzi, H.; and Zettlemoyer, L. 2020.
\newblock {A}mbig{QA}: Answering Ambiguous Open-domain Questions.
\newblock In Webber, B.; Cohn, T.; He, Y.; and Liu, Y., eds., \emph{Proceedings of the 2020 Conference on Empirical Methods in Natural Language Processing (EMNLP)}, 5783--5797. Online: Association for Computational Linguistics.

\bibitem[{Ning et~al.(2024)Ning, Su, Lv, Zhang, Liu, Liu, and Xu}]{ning2024wtuevalwhetherornottoolusage}
Ning, K.; Su, Y.; Lv, X.; Zhang, Y.; Liu, J.; Liu, K.; and Xu, J. 2024.
\newblock WTU-EVAL: A Whether-or-Not Tool Usage Evaluation Benchmark for Large Language Models.
\newblock arXiv:2407.12823.

\bibitem[{OpenAI(2023{\natexlab{a}})}]{openai2023chatgpt}
OpenAI. 2023{\natexlab{a}}.
\newblock ChatGPT: A Large Language Model developed by OpenAI.
\newblock \url{https://www.openai.com/chatgpt}.
\newblock Accessed: 2024-06-16.

\bibitem[{OpenAI(2023{\natexlab{b}})}]{few-shot-prompts}
OpenAI. 2023{\natexlab{b}}.
\newblock Prompt Engineering.
\newblock \url{https://platform.openai.com/docs/guides/prompt-engineering}.
\newblock Accessed: 2024-06-16.

\bibitem[{Ouyang et~al.(2024)Ouyang, Zhang, Yan, Liu, Choi, Han, and Qin}]{ouyang2024structured}
Ouyang, S.; Zhang, Z.; Yan, B.; Liu, X.; Choi, Y.; Han, J.; and Qin, L. 2024.
\newblock Structured chemistry reasoning with large language models.
\newblock In \emph{Proceedings of the 41st International Conference on Machine Learning}, 38937--38952.

\bibitem[{Patil et~al.(2025)Patil, Mao, Cheng-Jie~Ji, Yan, Suresh, Stoica, and E.~Gonzalez}]{patil2025bfcl}
Patil, S.~G.; Mao, H.; Cheng-Jie~Ji, C.; Yan, F.; Suresh, V.; Stoica, I.; and E.~Gonzalez, J. 2025.
\newblock The Berkeley Function Calling Leaderboard (BFCL): From Tool Use to Agentic Evaluation of Large Language Models.
\newblock In \emph{Forty-second International Conference on Machine Learning}.

\bibitem[{Patil et~al.(2024)Patil, Zhang, Wang, and Gonzalez}]{patil2023gorilla}
Patil, S.~G.; Zhang, T.; Wang, X.; and Gonzalez, J.~E. 2024.
\newblock Gorilla: Large language model connected with massive apis.
\newblock \emph{Advances in Neural Information Processing Systems}, 37: 126544--126565.

\bibitem[{Press et~al.(2023)Press, Zhang, Min, Schmidt, Smith, and Lewis}]{press2023measuring}
Press, O.; Zhang, M.; Min, S.; Schmidt, L.; Smith, N.~A.; and Lewis, M. 2023.
\newblock Measuring and Narrowing the Compositionality Gap in Language Models.
\newblock In \emph{Findings of the Association for Computational Linguistics: EMNLP 2023}, 5687--5711.

\bibitem[{Qin et~al.(2024)Qin, Liang, Ye, Zhu, Yan, Lu, Lin, Cong, Tang, Qian et~al.}]{qin2023toolllm}
Qin, Y.; Liang, S.; Ye, Y.; Zhu, K.; Yan, L.; Lu, Y.; Lin, Y.; Cong, X.; Tang, X.; Qian, B.; et~al. 2024.
\newblock ToolLLM: Facilitating Large Language Models to Master 16000+ Real-world APIs.
\newblock In \emph{The Twelfth International Conference on Learning Representations}.

\bibitem[{Ross, Mahabaleshwarkar, and Suhara(2025)}]{ross-etal-2025-when2call}
Ross, H.; Mahabaleshwarkar, A.~S.; and Suhara, Y. 2025.
\newblock {W}hen2{C}all: When (not) to Call Tools.
\newblock In Chiruzzo, L.; Ritter, A.; and Wang, L., eds., \emph{Proceedings of the 2025 Conference of the Nations of the Americas Chapter of the Association for Computational Linguistics: Human Language Technologies (Volume 1: Long Papers)}, 3391--3409. Albuquerque, New Mexico: Association for Computational Linguistics.
\newblock ISBN 979-8-89176-189-6.

\bibitem[{Schick et~al.(2024)Schick, Dwivedi-Yu, Dess{\`\i}, Raileanu, Lomeli, Hambro, Zettlemoyer, Cancedda, and Scialom}]{schick2024toolformer}
Schick, T.; Dwivedi-Yu, J.; Dess{\`\i}, R.; Raileanu, R.; Lomeli, M.; Hambro, E.; Zettlemoyer, L.; Cancedda, N.; and Scialom, T. 2024.
\newblock Toolformer: Language models can teach themselves to use tools.
\newblock \emph{Advances in Neural Information Processing Systems}, 36.

\bibitem[{Shinn et~al.(2023)Shinn, Cassano, Gopinath, Narasimhan, and Yao}]{shinn2023reflexion}
Shinn, N.; Cassano, F.; Gopinath, A.; Narasimhan, K.; and Yao, S. 2023.
\newblock Reflexion: Language agents with verbal reinforcement learning.
\newblock \emph{Advances in Neural Information Processing Systems}, 36: 8634--8652.

\bibitem[{Sun et~al.(2024)Sun, Min, Chang, and Bisk}]{sun-etal-2024-tools}
Sun, J.; Min, S.~Y.; Chang, Y.; and Bisk, Y. 2024.
\newblock Tools Fail: Detecting Silent Errors in Faulty Tools.
\newblock In \emph{Proceedings of the 2024 Conference on Empirical Methods in Natural Language Processing}, 14272--14289.

\bibitem[{Team et~al.(2025)Team, Kamath, Ferret, Pathak, Vieillard, Merhej, Perrin, Matejovicova, Ram{\'e}, Rivi{\`e}re et~al.}]{team2025gemma}
Team, G.; Kamath, A.; Ferret, J.; Pathak, S.; Vieillard, N.; Merhej, R.; Perrin, S.; Matejovicova, T.; Ram{\'e}, A.; Rivi{\`e}re, M.; et~al. 2025.
\newblock Gemma 3 technical report.
\newblock \emph{arXiv preprint arXiv:2503.19786}.

\bibitem[{Touvron et~al.(2023)Touvron, Martin, Stone, Albert, Almahairi, Babaei, Bashlykov, Batra, Bhargava, Bhosale et~al.}]{touvron2023llama}
Touvron, H.; Martin, L.; Stone, K.; Albert, P.; Almahairi, A.; Babaei, Y.; Bashlykov, N.; Batra, S.; Bhargava, P.; Bhosale, S.; et~al. 2023.
\newblock Llama 2: Open foundation and fine-tuned chat models.
\newblock \emph{arXiv preprint arXiv:2307.09288}.

\bibitem[{Wang et~al.(2024)Wang, Cheng, Zhu, Fried, and Neubig}]{tool-anyway}
Wang, Z.; Cheng, Z.; Zhu, H.; Fried, D.; and Neubig, G. 2024.
\newblock What Are Tools Anyway? A Survey from the Language Model Perspective.
\newblock In \emph{First Conference on Language Modeling}.

\bibitem[{Wolf et~al.(2020)Wolf, Debut, Sanh, Chaumond, Delangue, Moi, Cistac, Rault, Louf, Funtowicz, Davison, Shleifer, von Platen, Ma, Jernite, Plu, Xu, Scao, Gugger, Drame, Lhoest, and Rush}]{wolf-etal-2020-transformers}
Wolf, T.; Debut, L.; Sanh, V.; Chaumond, J.; Delangue, C.; Moi, A.; Cistac, P.; Rault, T.; Louf, R.; Funtowicz, M.; Davison, J.; Shleifer, S.; von Platen, P.; Ma, C.; Jernite, Y.; Plu, J.; Xu, C.; Scao, T.~L.; Gugger, S.; Drame, M.; Lhoest, Q.; and Rush, A.~M. 2020.
\newblock Transformers: State-of-the-Art Natural Language Processing.
\newblock In \emph{Proceedings of the 2020 Conference on Empirical Methods in Natural Language Processing: System Demonstrations}, 38--45. Online: Association for Computational Linguistics.

\bibitem[{Xu et~al.(2023)Xu, Hong, Li, Hu, Chen, and Zhang}]{xu2023tool}
Xu, Q.; Hong, F.; Li, B.; Hu, C.; Chen, Z.; and Zhang, J. 2023.
\newblock On the Tool Manipulation Capability of Open-sourced Large Language Models.
\newblock In \emph{NeurIPS 2023 Foundation Models for Decision Making Workshop}.

\bibitem[{Yan et~al.(2024)Yan, Mao, Ji, Zhang, Patil, Stoica, and Gonzalez}]{berkeley-function-calling-leaderboard}
Yan, F.; Mao, H.; Ji, C. C.-J.; Zhang, T.; Patil, S.~G.; Stoica, I.; and Gonzalez, J.~E. 2024.
\newblock Berkeley Function Calling Leaderboard.
\newblock \url{https://gorilla.cs.berkeley.edu/blogs/8_berkeley_function_calling_leaderboard.html}.

\bibitem[{Yang et~al.(2024)Yang, Yang, Zhang, Hui, Zheng, Yu, Li, Liu, Huang, Wei et~al.}]{yang2024qwen2.5}
Yang, A.; Yang, B.; Zhang, B.; Hui, B.; Zheng, B.; Yu, B.; Li, C.; Liu, D.; Huang, F.; Wei, H.; et~al. 2024.
\newblock Qwen2. 5 Technical Report.
\newblock \emph{arXiv preprint arXiv:2412.15115}.

\bibitem[{Yang, Yue, and He(2023)}]{autogpt}
Yang, H.; Yue, S.; and He, Y. 2023.
\newblock Auto-gpt for online decision making: Benchmarks and additional opinions.
\newblock \emph{arXiv preprint arXiv:2306.02224}.

\bibitem[{Yao et~al.(2023)Yao, Zhao, Yu, Du, Shafran, Narasimhan, and Cao}]{react}
Yao, S.; Zhao, J.; Yu, D.; Du, N.; Shafran, I.; Narasimhan, K.~R.; and Cao, Y. 2023.
\newblock ReAct: Synergizing Reasoning and Acting in Language Models.
\newblock In \emph{The Eleventh International Conference on Learning Representations}.

\bibitem[{Ye et~al.(2024{\natexlab{a}})Ye, Li, Li, Huang, Gao, Wu, Zhang, Gui, and Huang}]{ye-etal-2024-toolsword}
Ye, J.; Li, S.; Li, G.; Huang, C.; Gao, S.; Wu, Y.; Zhang, Q.; Gui, T.; and Huang, X. 2024{\natexlab{a}}.
\newblock {T}ool{S}word: Unveiling Safety Issues of Large Language Models in Tool Learning Across Three Stages.
\newblock In Ku, L.-W.; Martins, A.; and Srikumar, V., eds., \emph{Proceedings of the 62nd Annual Meeting of the Association for Computational Linguistics (Volume 1: Long Papers)}, 2181--2211. Bangkok, Thailand: Association for Computational Linguistics.

\bibitem[{Ye et~al.(2024{\natexlab{b}})Ye, Wu, Gao, Huang, Li, Li, Fan, Zhang, Gui, and Huang}]{ye2024rotbench}
Ye, J.; Wu, Y.; Gao, S.; Huang, C.; Li, S.; Li, G.; Fan, X.; Zhang, Q.; Gui, T.; and Huang, X.-J. 2024{\natexlab{b}}.
\newblock RoTBench: A Multi-Level Benchmark for Evaluating the Robustness of Large Language Models in Tool Learning.
\newblock In \emph{Proceedings of the 2024 Conference on Empirical Methods in Natural Language Processing}, 313--333.

\bibitem[{Yuan et~al.(2025)Yuan, Song, Chen, Tan, Shen, Ren, Li, and Yang}]{yuan2024easytool}
Yuan, S.; Song, K.; Chen, J.; Tan, X.; Shen, Y.; Ren, K.; Li, D.; and Yang, D. 2025.
\newblock EASYTOOL: Enhancing LLM-based Agents with Concise Tool Instruction.
\newblock In \emph{Proceedings of the 2025 Conference of the Nations of the Americas Chapter of the Association for Computational Linguistics: Human Language Technologies (Volume 1: Long Papers)}, 951--972.

\bibitem[{Zhang et~al.(2024{\natexlab{a}})Zhang, Li, Li, Shi, and Jin}]{zhang-etal-2024-codeagent}
Zhang, K.; Li, J.; Li, G.; Shi, X.; and Jin, Z. 2024{\natexlab{a}}.
\newblock {C}ode{A}gent: Enhancing Code Generation with Tool-Integrated Agent Systems for Real-World Repo-level Coding Challenges.
\newblock In Ku, L.-W.; Martins, A.; and Srikumar, V., eds., \emph{Proceedings of the 62nd Annual Meeting of the Association for Computational Linguistics (Volume 1: Long Papers)}, 13643--13658. Bangkok, Thailand: Association for Computational Linguistics.

\bibitem[{Zhang et~al.(2024{\natexlab{b}})Zhang, Chen, Wang, Liu, Yang, Shi, Zhu, Lin, Wan, Yang, Sakai, Feng, and Yamana}]{zhang-etal-2024-toolbehonest}
Zhang, Y.; Chen, J.; Wang, J.; Liu, Y.; Yang, C.; Shi, C.; Zhu, X.; Lin, Z.; Wan, H.; Yang, Y.; Sakai, T.; Feng, T.; and Yamana, H. 2024{\natexlab{b}}.
\newblock {T}ool{B}e{H}onest: A Multi-level Hallucination Diagnostic Benchmark for Tool-Augmented Large Language Models.
\newblock In Al-Onaizan, Y.; Bansal, M.; and Chen, Y.-N., eds., \emph{Proceedings of the 2024 Conference on Empirical Methods in Natural Language Processing}, 11388--11422. Miami, Florida, USA: Association for Computational Linguistics.

\bibitem[{Zheng et~al.(2024)Zheng, Chiang, Sheng, Zhuang, Wu, Zhuang, Lin, Li, Li, Xing et~al.}]{zheng2024judging}
Zheng, L.; Chiang, W.-L.; Sheng, Y.; Zhuang, S.; Wu, Z.; Zhuang, Y.; Lin, Z.; Li, Z.; Li, D.; Xing, E.; et~al. 2024.
\newblock Judging llm-as-a-judge with mt-bench and chatbot arena.
\newblock \emph{Advances in Neural Information Processing Systems}, 36.

\end{thebibliography}

\section{Supplementary Material}
This supplementary material includes limitations and future work, a detailed dataset verification process, further analysis, generated data samples, and prompt templates used in our overall experiments, and  the reproducibility checklist. The source code and data will be made publicly available.

\section{Limitations and Future Work}
Our study contains several limitations. The data annotation relies on model-based annotations, with human verification, which may differ from real-world situations where humans provide incomplete information. Furthermore, our focus on API-based tools does not cover the broader spectrum of tools, such as plugins and robotic systems. Nevertheless, our research highlights the importance of developing reliable tool-augmented LLMs and points to the need for continued advancements in this area to improve model robustness and safety in real-world applications.

\section{Detailed Dataset Verification Process}
Our preliminary analysis revealed that automatic generation sometimes failed to create truly incomplete scenarios, making manual verification crucial for maintaining dataset quality. 
To resolve this, we conducted the dataset verification process involved rigorous criteria for both manipulation types. For \textit{Relevant API Replacement}, we removed instances where:
\begin{itemize}
\item The user's request could be fulfilled using alternative APIs
\item The request could be handled using non-replaced APIs, particularly in ToolBench instances
\end{itemize}
For \textit{Partial Removal of User Utterance}, we filtered out cases where:
\begin{itemize}
\item Only non-essential information was removed, allowing APIs to still fulfill the request
\item One of multiple requests was removed, but the remaining requests could be resolved through existing API invocations
\item The conversation context became unnatural or grammatically incorrect
\item The result was identical to the original instance
\item The last utterance no longer pertained to the user, making API execution unnatural
\end{itemize}
Two authors independently reviewed the instances, discussing and resolving any disagreements to ensure consistent application of these criteria. This process was iteratively refined until we achieved a high level of agreement on the identification of truly incomplete scenarios.

% \usepackage{multirow}
\begin{table*}[ht!]
\centering
\setlength{\tabcolsep}{1mm}
% \begin{adjustbox}{width=\textwidth}
\begin{tabular}{ccccccccccccc}
\toprule
 &  &  & \multicolumn{1}{c}{\begin{tabular}[c]{@{}c@{}}Llama\\ 3.1 8B\end{tabular}} & \multicolumn{1}{c}{\begin{tabular}[c]{@{}c@{}}Llama\\ 3.1 70B\end{tabular}} & 

 \multicolumn{1}{c}{\begin{tabular}[c]{@{}c@{}}Gemma\\ 3-4B\end{tabular}} & 
 \multicolumn{1}{c}{\begin{tabular}[c]{@{}c@{}}Gemma\\ 3-27B\end{tabular}} & 
 \multicolumn{1}{c}{\begin{tabular}[c]{@{}c@{}}Qwen\\ 2.5-7B\end{tabular}} & 
 \multicolumn{1}{c}{\begin{tabular}[c]{@{}c@{}}Qwen\\ 2.5-14B\end{tabular}} & 
 \multicolumn{1}{c}{\begin{tabular}[c]{@{}c@{}}Qwen\\ 2.5-32B\end{tabular}} & 
 \multicolumn{1}{c}{\begin{tabular}[c]{@{}c@{}}Qwen\\ 2.5-72B\end{tabular}} & \multicolumn{1}{c}{\begin{tabular}[c]{@{}c@{}}GPT-4o\\ mini\end{tabular}} & \multicolumn{1}{c}{GPT-4o} \\
 \hline
 
 \multicolumn{9}{l}{\textit{\textbf{API Replacement}}} \\
 \hline

\multirow{6}{*}{APIBank} & \multirow{2}{*}{Acc.} & 0-shot  & 76.06 & 79.10 & 67.85 & 78.49 & 73.05 & 80.97 & 85.82 & 87.94 & 88.82 & 86.03 \\
 &  & CoT  & 78.13 & 78.61 & 70.33 & 79.79 & 76.12 & 85.11 & 86.05 & 85.46 &82.38 & 82.99 \\
 & & Ours  &	76.24 &	84.52 &	71.51 & 77.90 & 77.78 & 84.99 & 84.52 & 88.53 &84.75 &	85.58 \\ \cline{2-13}
 & \multirow{2}{*}{F1} & 0-shot & 71.32 & 81.22 & 57.10 & 73.31 & 65.97 & 77.35 & 83.83 & 87.02 &88.41 & 86.23 \\ 
 &  & CoT & 78.92 & 81.47& 59.19 & 75.54 & 72.63 & 84.13 & 84.95 & 85.16 &83.50 & 84.38 \\
 &  & Ours&78.50&84.61&68.90 & 75.75 & 77.73 & 84.61 & 84.24 & 88.49 &85.29&86.04 \\ \cline{2-13}
\multirow{6}{*}{ToolBench} & \multirow{2}{*}{Acc.} & 0-shot & 69.45 & 72.64 &  55.56 & 64.36 & 72.33 & 69.29 & 70.55 & 74.11 & 77.69 & 76.48 \\
 &  & CoT & 75.16 & 64.95 &59.54 & 70.13 & 76.21 & 74.53 & 76.73 & 80.50 & 64.51 & 70.33 \\
 & & Ours  &	67.71 &	70.65  &65.41 & 70.75 & 70.75 & 72.85 & 78.93 & 83.12 &	69.60 &	71.07 \\ \cline{2-13}
 & \multirow{2}{*}{F1} & 0-shot  & 56.43 & 76.97  &20.90 & 45.51 & 64.80 & 57.96 & 58.74 & 67.80 & 72.97 & 75.06 \\
 &  & CoT  & 77.40 & 72.80 & 33.68 & 58.87 & 75.41 & 66.48 & 77.16 & 80.38 &70.87 & 75.27 \\
 &  & Ours&72.35&74.31&68.57 & 71.96 & 73.50 & 75.82 & 75.93 & 81.89 &74.34&75.92 \\
 \hline
 \multicolumn{9}{l}{\textit{\textbf{Utterance Removal}}} \\ \hline
\multirow{6}{*}{APIBank} & \multirow{2}{*}{Acc.} & 0-shot  & 52.63 & 77.42 & 53.12 & 62.66 & 54.77 & 79.28 & 81.09 & 81.25 & 77.25 & 88.29   \\
 &  & CoT  & 61.80 & 81.66  & 54.28 & 67.93 & 69.90 & 82.73 & 86.35 & 88.49 &86.59 & 85.74   \\
 & & Ours &	76.48 &	87.34  &78.45 & 88.98 & 83.06 & 89.97 & 89.80 & 91.12 &	87.50 &	89.97 \\ \cline{2-13}
 & \multirow{2}{*}{F1} & 0-shot  & 19.60 & 80.06 &21.49 & 42.24 & 26.67 & 74.80 & 77.76 & 77.29 & 73.41 & 88.74   \\
 &  & CoT & 55.97 & 83.98 &18.24 & 54.55 & 62.11 & 81.61 & 85.26 & 88.41 & 86.72 & 86.96   \\
 &  &Ours &78.43&87.91&78.35 & 89.21 & 83.93 & 90.18 & 90.31 & 91.43 &88.31&90.72 \\ \cline{2-13}
\multirow{6}{*}{ToolBench} & \multirow{2}{*}{Acc.} & 0-shot  & 50.19 & 71.54 & 50.00 & 52.09 & 50.86 & 50.99 & 54.68 & 61.45 & 57.57 & 74.51   \\
 &  & CoT  & 52.26 & 68.31  &50.86 & 53.20 & 59.85 & 57.14 & 67.73 & 68.35 & 58.99 & 72.19   \\
 & & Ours  &	62.32 &	74.14&71.92 & 77.22 & 70.20 & 76.23 & 79.80 & 80.79 &	75.86 &	77.09 \\ \cline{2-13}
 & \multirow{2}{*}{F1} & 0-shot  & 5.41 & 74.94  &1.46 & 10.16 & 5.23 & 15.68 & 21.37 & 43.19 & 35.18 & 70.90   \\
 &  & CoT  & 45.17 & 74.56 &6.12 & 18.45 & 39.85 & 43.14 & 63.91 & 62.26 &64.82 & 75.87  \\ 
 & & Ours &65.15&76.77 &74.55 & 78.16 & 70.77 & 77.69 & 81.36 & 81.16 &79.10&80.17 \\
 \bottomrule

\end{tabular}
% \end{adjustbox}
\caption{Performance evaluation results of LLM by manipulation type. The accuracy (Acc.) and F1 score (F1) are used for evaluation metrics. Both the zero-shot, Chain-of-Thought (CoT), and Structured verification (Ours) performance are presented. We used an instruction-tuned version of open-source models for the overall experiments.}
\label{tab:cot_full}
\end{table*}

\section{Full results of Structured Verification}
Table \ref{tab:cot_full} presents the evaluation results of different models with different prompting strategies (i.e., zero-shot, CoT, and Structured verification (Ours)).

\section{Further Analysis}

We perform analyses to understand how LLMs perceive and respond in incomplete conditions.

\begin{table}[t!]
\setlength{\tabcolsep}{1mm}
% \begin{adjustbox}{width=\columnwidth}
\centering
\begin{tabular}{lcccccc}
\toprule
 &\textbf{7B}$_{\textsc{base}}$ & \textbf{7B} & \textbf{13B}$_{\textsc{base}}$ & \textbf{13B} & \textbf{70B}$_{\textsc{base}}$ & \textbf{70B}  \\ \hline
     
\multicolumn{7}{l}{\textit{\textbf{API Replacement}}} \\ \hline
Acc. & 50.06 & 49.94 & 54.86 & 53.75 & \textbf{60.52} & \underline{59.66} \\
F1 & 0 & 0.94 & \textbf{64.05} & 14.55 & \underline{56.68} & 36.19 \\ \hline
\multicolumn{7}{l}{\textit{\textbf{Utterance Removal}}} \\ \hline
Acc. & 50.00 & 50.49 & \underline{53.01} & 51.29 & \textbf{54.91} & 52.15 \\
F1 & 0 & 3.22 & \textbf{62.25} & 5.98 & \underline{45.87} & 16.77 \\ \hline
\end{tabular}
% \end{adjustbox}
\caption{Performance comparison between Llama-2 model family \citep{touvron2023llama} of different sizes and training approaches. Models not trained on instruction-tuning datasets are marked as \textsc{BASE}.}
\label{tab:llama-size}
\end{table}

\subsection{Impacts of Model Size and Instruction Tuning}
We analyze the impact of model size and instruction-tuning on model performance across our dataset. Specifically, we compare Llama-2-7B, Llama-2-13B, and Llama-2-70B models against their instruction-tuned counterparts, Llama-2-7B-chat, Llama-2-13B-chat, and Llama-2-70B-chat. The evaluation results, presented in Table~\ref{tab:llama-size}, reveal that both the 7B and 7B-chat models frequently predict Yes for most instances, resulting in very low F1 scores. Overall, models that were not trained on instruction-following data generally outperform their instruction-tuned counterparts, regardless of model size. Interestingly, while instruction-tuned models show performance improvements as the model size increases, the non-instruction-tuned models exhibit varying trends, with a decrease in F1 score as the model size scales from 13B to 70B. These findings suggest that training on instruction-following datasets may affect the models' ability to recognize incomplete conditions, indicating a need for further exploration in future work.

% \begin{table}[t]
% \setlength{\tabcolsep}{1mm}
% \begin{adjustbox}{width=\columnwidth}
% \centering
% \begin{tabular}{lcccc}
% \toprule
% \multicolumn{1}{c}{} & \textbf{Mistral-7B} & \textbf{Claude-3} & \textbf{GPT-3.5-T} & \textbf{GPT-4} \\
% \hline
% \multicolumn{5}{l}{\textit{\textbf{API Replacement}}}                      \\ \hline
% \textbf{Yes Ratio}   & 87.74/88.79         & 70.69/91.91       & 81.37/92.74      & 75.74/57.25    \\ \hline
% \multicolumn{5}{l}{\textit{\textbf{Utterance Removal}}}                      \\ \hline
% \textbf{Yes Ratio}   & 97.16/98.22         & 84.41/99.51       & 97.65/99.13      & 91.09/73.33   \\

% \bottomrule
% \end{tabular}
% \end{adjustbox}
% \caption{Predictive distribution on main experiments of ToolBench. The indicators are the same as Table \ref{tab:appendix_table_yes_no_APIBank}.}
% \label{tab:appendix_table_yes_no_ToolBench}
% \end{table}

\begin{figure*}
\centering
\includegraphics[width=0.95\textwidth]{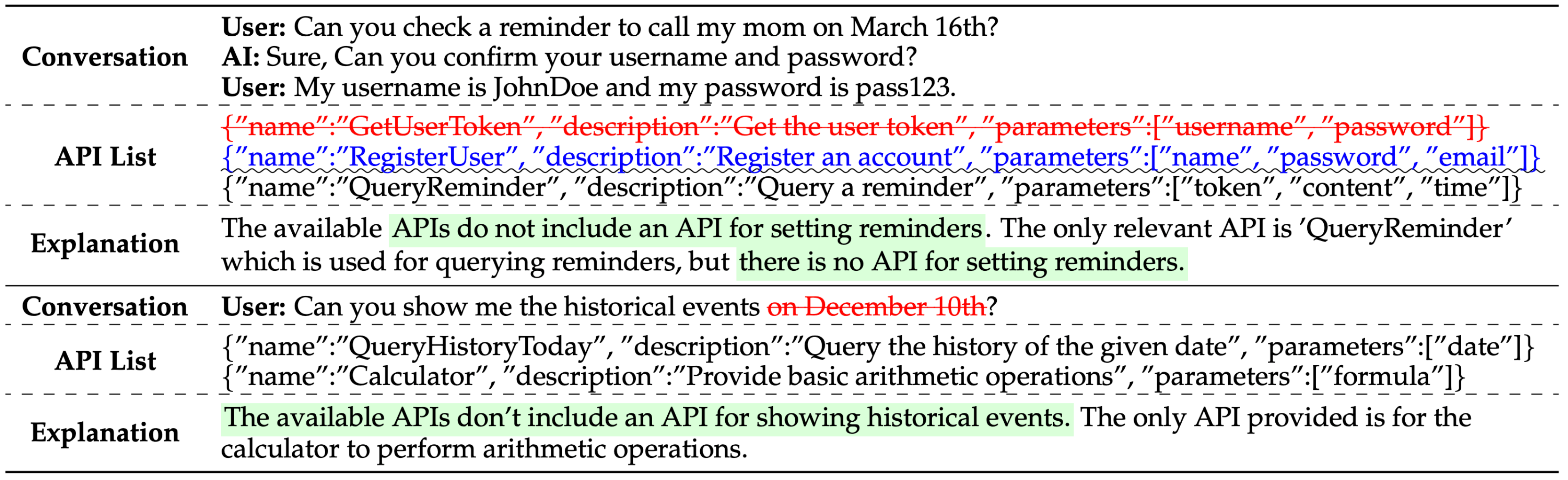}
  
\caption{\textbf{Examples of Incorrect Explanation}. The upper example illustrates a case of erroneous explanation in API Replacement, while the lower example shows one in Utterance Removal. 
Note that the model correctly identifies both instances as incomplete within a binary classification setup. 
Removed and newly included information, as part of our manipulation strategy, are highlighted with a strikethrough and wavy underline, respectively. Wrong explanations from the models are manually highlighted by the authors.}
\label{fig:case_study} 
\end{figure*}

\begin{table}[t!]
\centering
\begin{tabular}{lcccc}
\toprule
                    & \multicolumn{2}{c}{\textbf{API Replacement}} & \multicolumn{2}{c}{\textbf{Utterance Removal}} \\
                    & Num.                    & Acc. & Num.                       & Acc.                       \\
                    \hline

\noalign{\vspace{1.3ex}}                
% Mistral-7B & 149                              & 87.25                       & 19                                  & 47.37                          \\
GPT-3.5-T       & 182                              & 85.16                       & 52                                  & 48.08                          \\
 GPT-4               & 360                              & 97.50                        & 273                                 & 99.63      \\
\bottomrule
\end{tabular}
% \end{adjustbox}

\caption{The results of the explanation for the incomplete scenario. 
\textit{Num.} represents the count of accurately identified incomplete instances, which corresponds to the number of instances evaluated in the explanation assessment. \textit{Acc.} denotes accuracy of explanations judged by GPT-4, respectively.
}
\label{tab:task_2}
\end{table}

\subsection{Can Tool-augmented LLMs Correctly Explain Incomplete Conditions?}

We probe whether LLMs can accurately explain their decision-making process when they correctly identify incomplete conditions. To this end, we instruct the models to generate explanations for their decisions and assess whether these explanations correctly identify why tools cannot be used. We adopt the Judge LLM (i.e., GPT-4)~\citep{zheng2024judging} to evaluate the correctness of the explanations. We manually craft four-shot examples with a balanced class distribution to ensure a reliable evaluation.

In Table \ref{tab:task_2}, we observe that GPT-3.5-Turbo achieves an accuracy of 85.16\% and  48.08\% for API Replacement and Utterance Removal, respectively.
% , while Mistral-7B shows a performance with an accuracy of 87.25\%. For Utterance Removal, the accuracy of GPT-3.5-Turbo and Mistral-7B are and 47.37\%, respectively, showing similar performance.
These results indicate that it is more challenging for LLMs to provide accurate explanations when users offer insufficient context (Utterance Removal) compared to when the necessary tools are unavailable (API Replacement).

To further verify the reliability of Judge LLM in assessing explanation validity, we manually annotated 100 randomly sampled instances from GPT-3.5-Turbo's predictions, evenly divided between API Replacement and Utterance Removal. The agreement rate between human evaluators and Judge LLM is 82\%. This high level of agreement suggests that Judge LLM's evaluations are closely aligned with those of human evaluators, establishing it as a credible and effective assessment tool.

Additionally, we examine instances where the LLMs generated incorrect reasoning, as shown in Fig. \ref{fig:case_study}. In API Replacement, LLMs often misunderstand the user's intent, leading to inaccurate assertions that the available APIs are insufficient. Conversely, in Utterance Removal, the predominant errors stem from incorrect explanations asserting that no appropriate APIs are available, even when they are present.

\section{Samples of Incomplete Instances and Explanation}

\begin{figure*}
\centering
\includegraphics[width=0.8\textwidth]{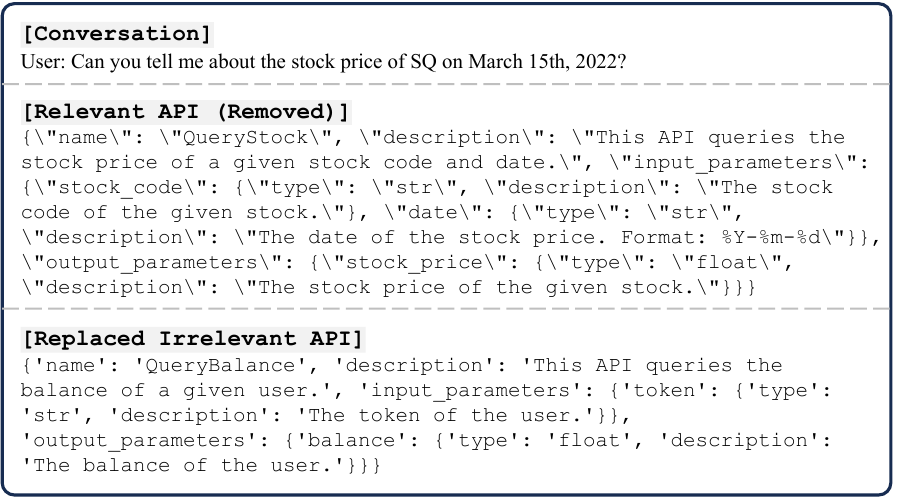}
  
\caption{\textit{API Replacement} Successful case from APIBank instances.}
\label{fig:app:apibank_api_replacement_success}
 
\end{figure*}

\begin{figure*}
\centering
\includegraphics[width=0.8\textwidth]{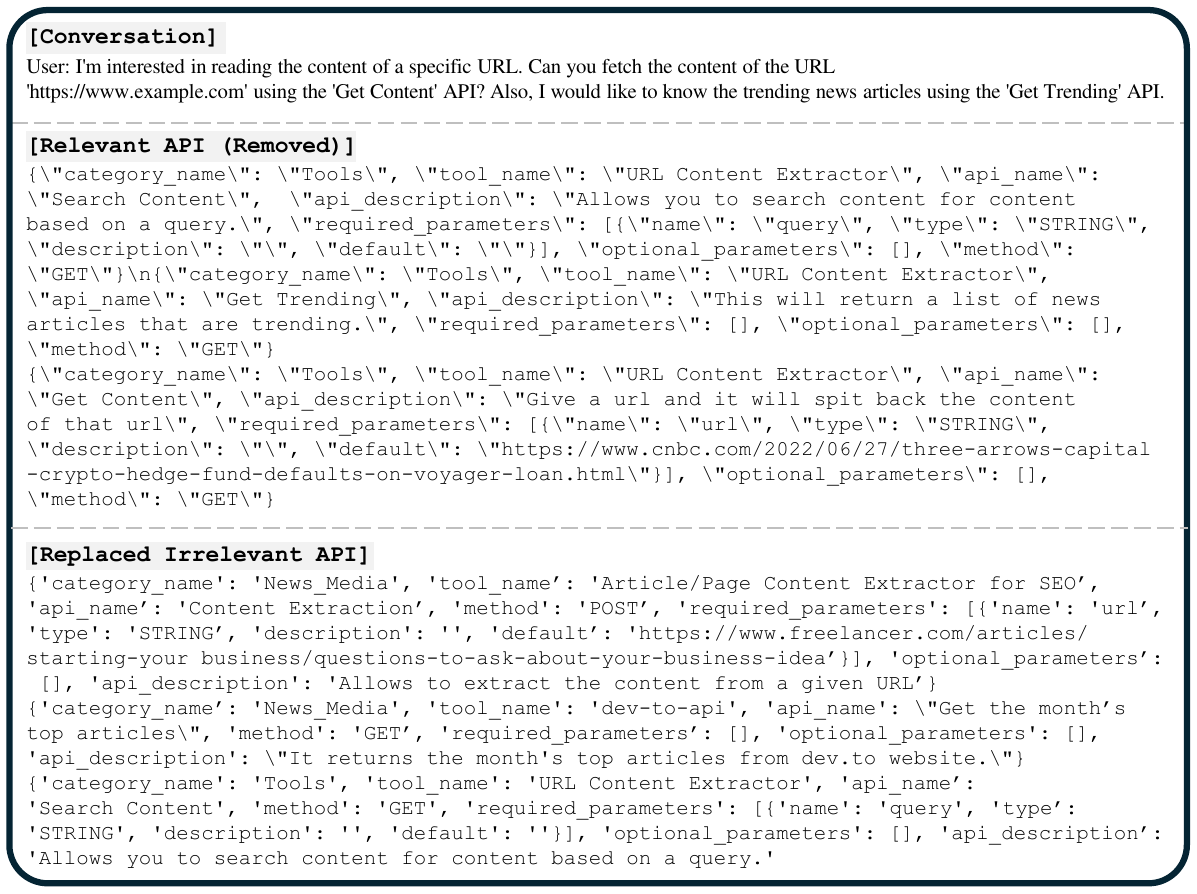}
  
\caption{\textit{API Replacement} Successful case from ToolBench instances.}
\label{fig:app:toolbench_api_replacement_success}
 
\end{figure*}

\begin{figure*}
\centering
\includegraphics[width=0.8\textwidth]{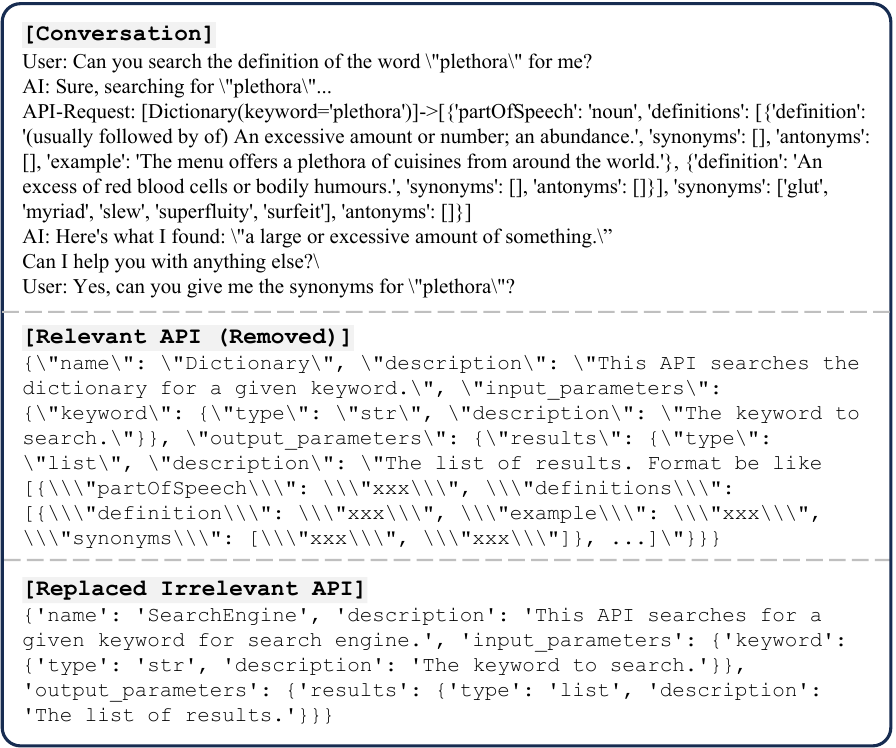}
  
\caption{\textit{API Replacement} Failed case from APIBank instances.}
\label{fig:app:apibank_api_replacement_wrong}
 
\end{figure*}

\begin{figure*}
\centering
\includegraphics[width=0.8\textwidth]{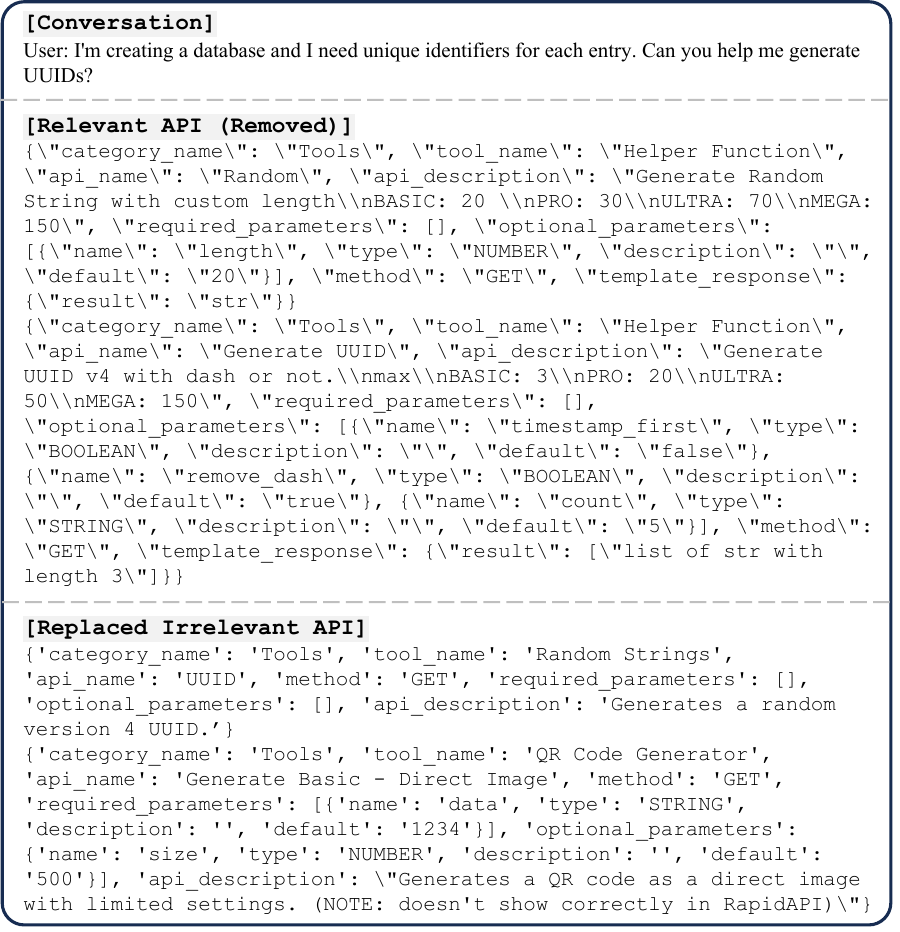}
  
\caption{\textit{API Replacement} Failed case from ToolBench instances.}
\label{fig:app:toolbench_api_replacement_wrong}
 
\end{figure*}

\begin{figure*}
\centering
\includegraphics[width=0.8\textwidth]{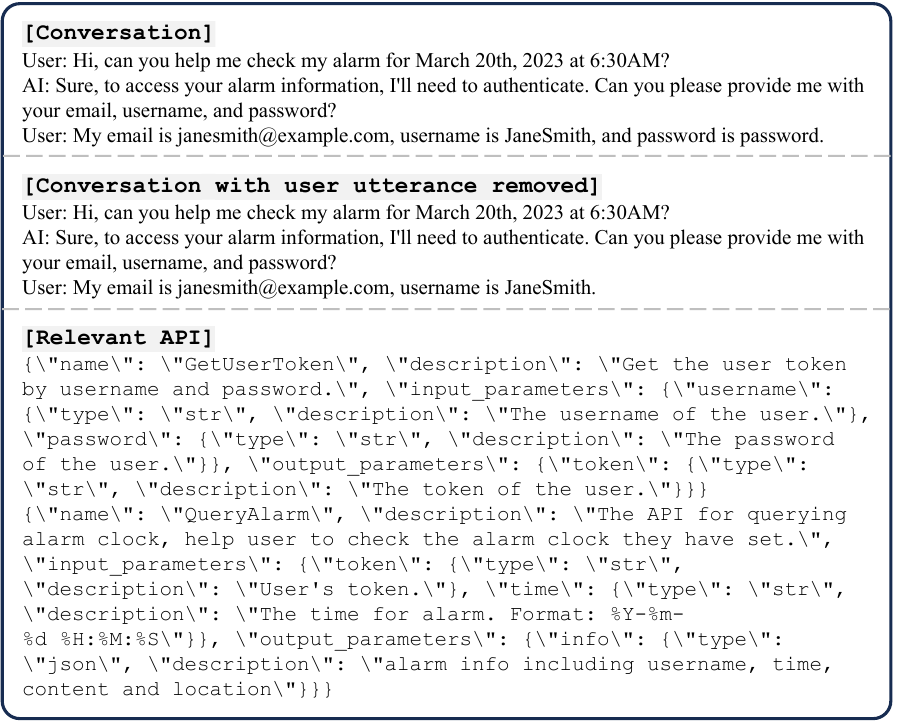}
  
\caption{ \textit{Utterance Removal} Successful case from APIBank instances.}
\label{fig:app:apibank_utterance_removal_success}
 
\end{figure*}

\begin{figure*}
\centering
\includegraphics[width=0.8\textwidth]{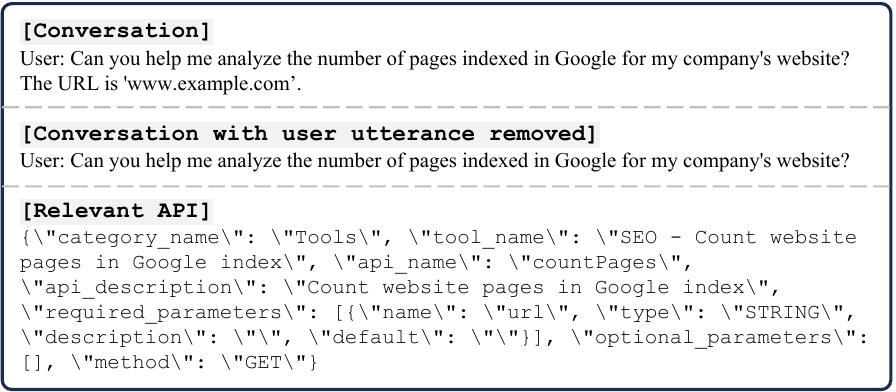}
  
\caption{\textit{Utterance Removal} Successful case from ToolBench instances.}
\label{fig:app:toolbench_utterance_removal_success}
 
\end{figure*}

\begin{figure*}
\centering
\includegraphics[width=0.8\textwidth]{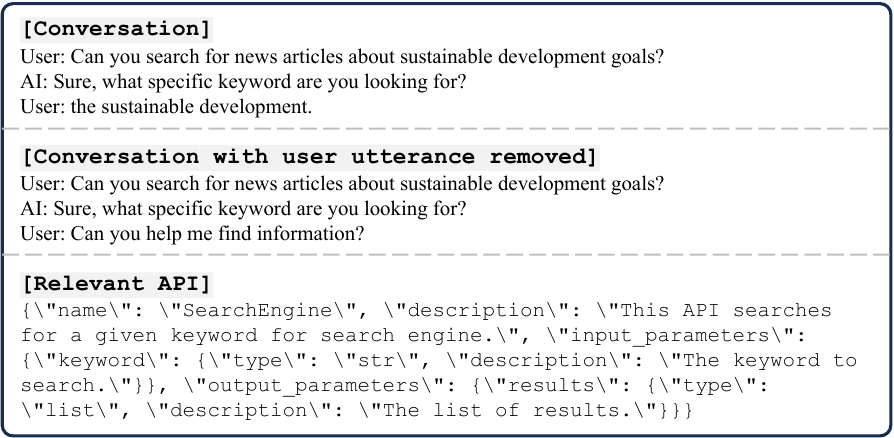}
  
\caption{\textit{Utterance Removal} Failed case from APIBank instances.}
\label{fig:app:apibank_utterance_removal_wrong}
 
\end{figure*}

\begin{figure*}
\centering
\includegraphics[width=0.8\textwidth]{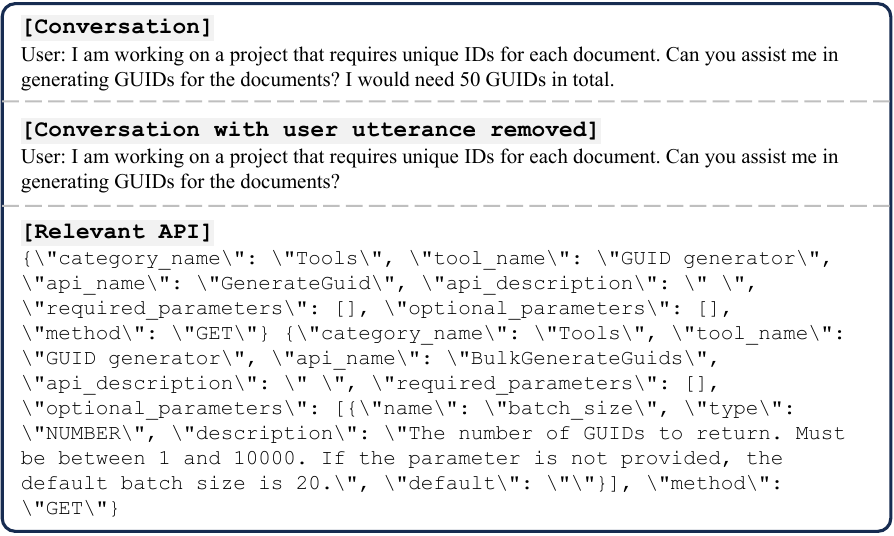}
  
\caption{\textit{Utterance Removal} Failed case from ToolBench instances.}
\label{fig:app:toolbench_utterance_removal_wrong}
 
\end{figure*}

\begin{figure*}
\centering
\includegraphics[width=0.8\textwidth]{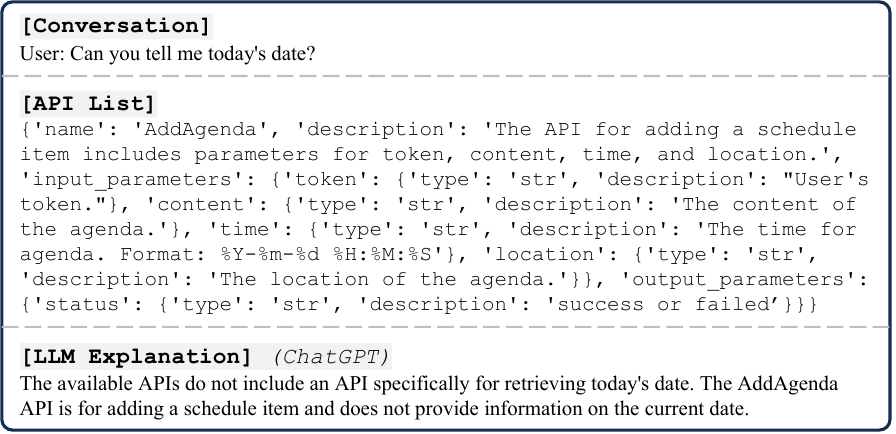}
  
\caption{Explanation Success Case in \textit{API Replacement}. The model clearly states and explains that an API to know current date is not provided.}
\label{fig:app:appendix_case_study_api_correct}
 
\end{figure*}

\begin{figure*}
\centering
\includegraphics[width=0.8\textwidth]{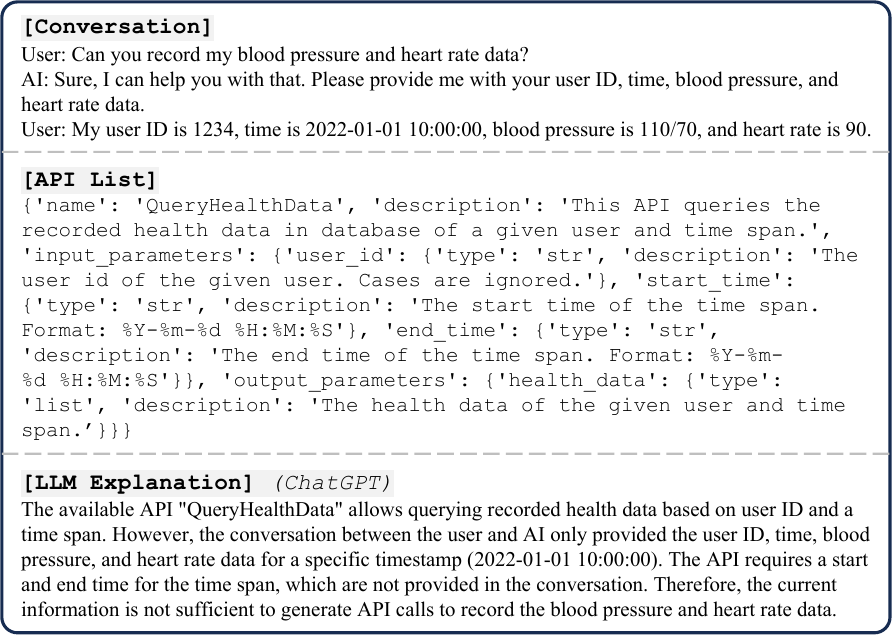}
  
\caption{Example of failed explanation in \textit{API Replacement}. The user wants to record health data, but only the \textit{QueryHealthData} API, which is for searching health data, is provided instead of an API for recording health data. The model incorrectly assumes that it must use the \textit{QueryHealthData} API and provides an erroneous explanation that it cannot call the API due to insufficient information.}
\label{fig:app:appendix_case_study_api_wrong}
 
\end{figure*}

\begin{figure*}
\centering
\includegraphics[width=0.8\textwidth]{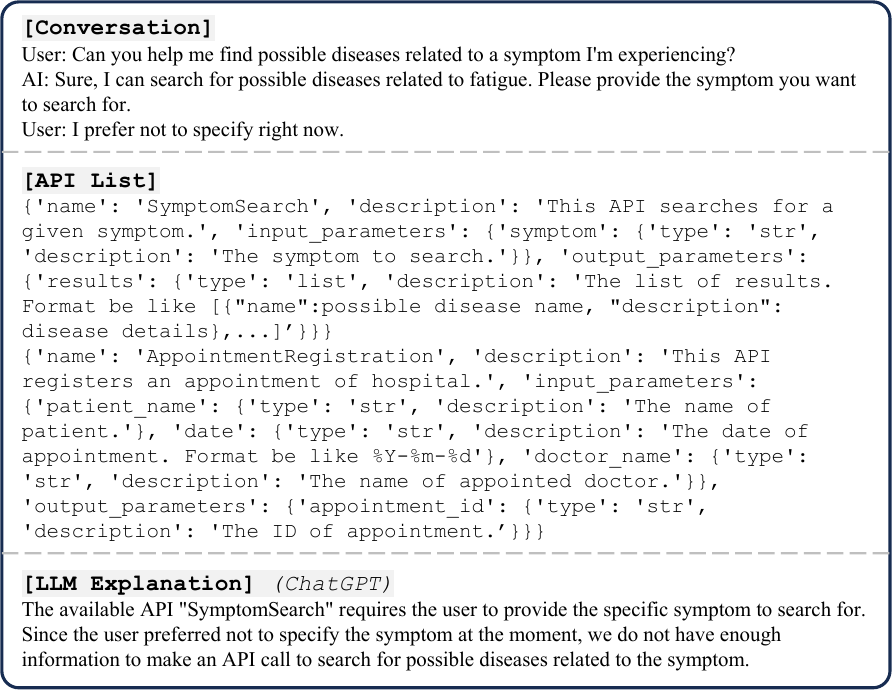}
  
\caption{Example of successful explanation in \textit{Utterance Removal}. The model recognizes the need to call the \textit{SymptomSearch} API to address the user's requirements. Additionally, it clearly acknowledges and explains that the necessary symptom information required to call the API has not been provided.}
\label{fig:app:appendix_case_study_user_correct}
 
\end{figure*}

\begin{figure*}
\centering
\includegraphics[width=0.8\textwidth]{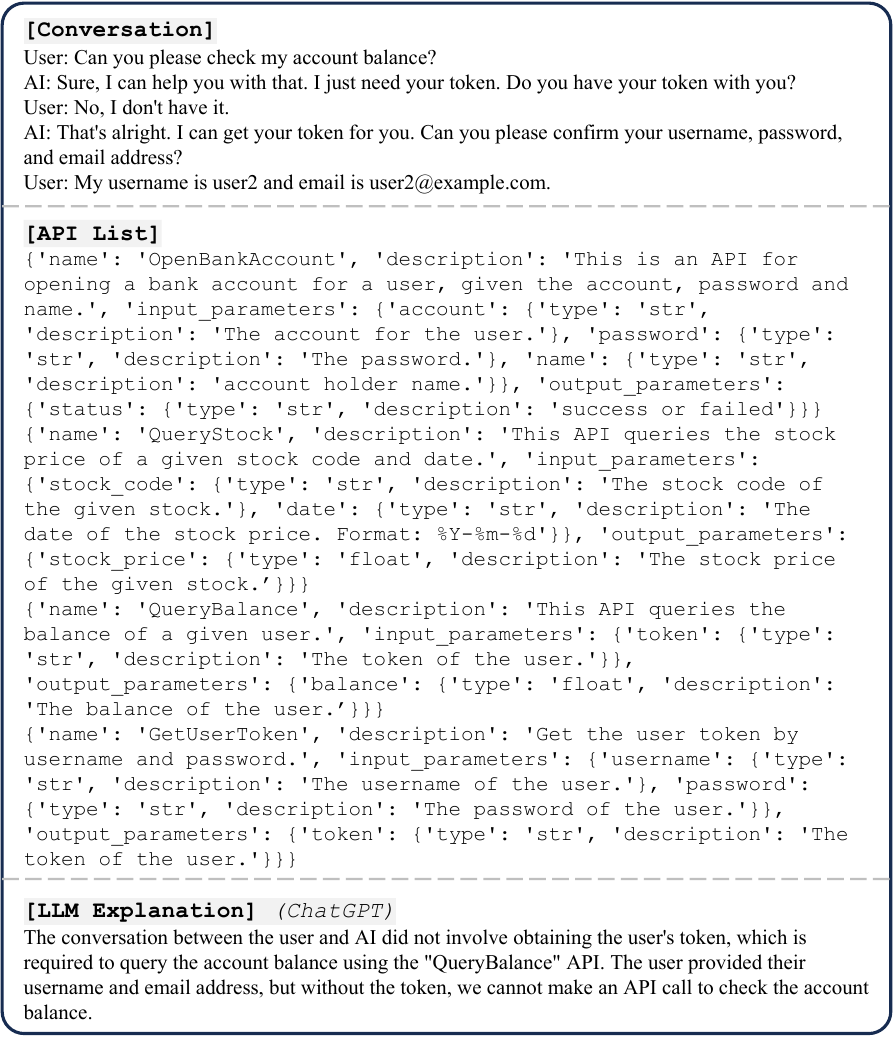}
  
\caption{Example of failed explanation in \textit{Utterance Removal}. The user's requirement is to check the account balance, which necessitates a token. However, the user has not provided the password required to obtain the token. Although an API exists to obtain the token, the model fails to recognize this and incorrectly states that the API call cannot be made due to the absence of the token.}
\label{fig:app:appendix_case_study_user_wrong}
 
\end{figure*}

We present the manipulated instances with different strategies are presented from 
Fig.~\ref{fig:app:apibank_api_replacement_success} to Fig.~\ref{fig:app:toolbench_utterance_removal_wrong}.
Both accurately and inaccurately modified instances resulting from our dataset construction method are provided. We also present additional explanation examples generated by LLMs when they recognize incomplete conditions on manipulated samples.
Examples of explanations in API Replacement are illustrated in 
Fig.~\ref{fig:app:appendix_case_study_api_correct} and Fig.~\ref{fig:app:appendix_case_study_api_wrong}.
Examples of explanations in Utterance Removal are depicted in 
Fig.~\ref{fig:app:appendix_case_study_user_correct} and 
Fig.~\ref{fig:app:appendix_case_study_user_wrong}.

\section{Prompt Templates}

\begin{figure*}
\centering
\includegraphics[width=0.8\textwidth]{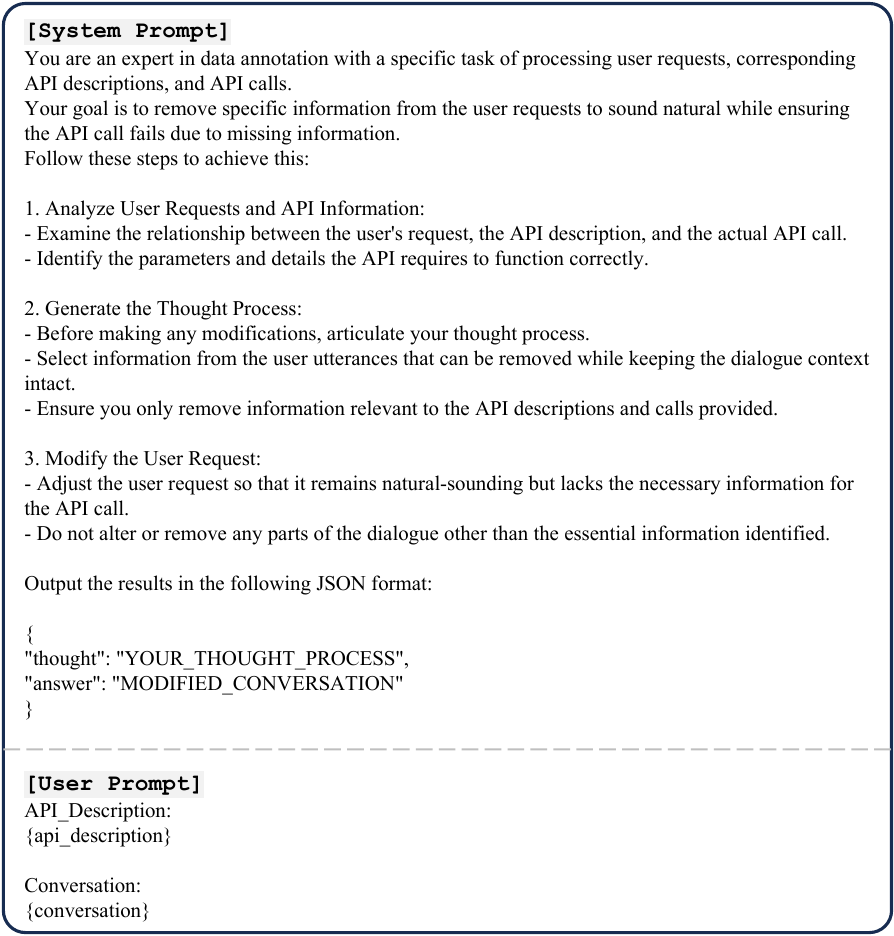}
  
\caption{Data Construction Prompt Template.}
\label{fig:app:data_construction_prompt_template}
 
\end{figure*}

\begin{figure*}
\centering
\includegraphics[width=0.8\textwidth]{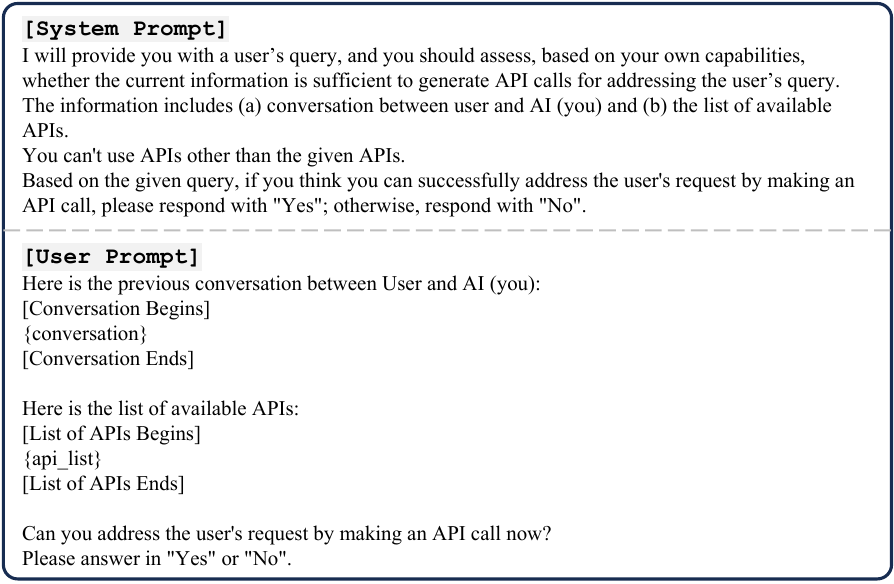}
  
\caption{Zero-shot Experiments Prompt Template.}
\label{fig:app:task1_prompt_template}
 
\end{figure*}

\begin{figure*}
\centering
\includegraphics[width=0.8\textwidth]{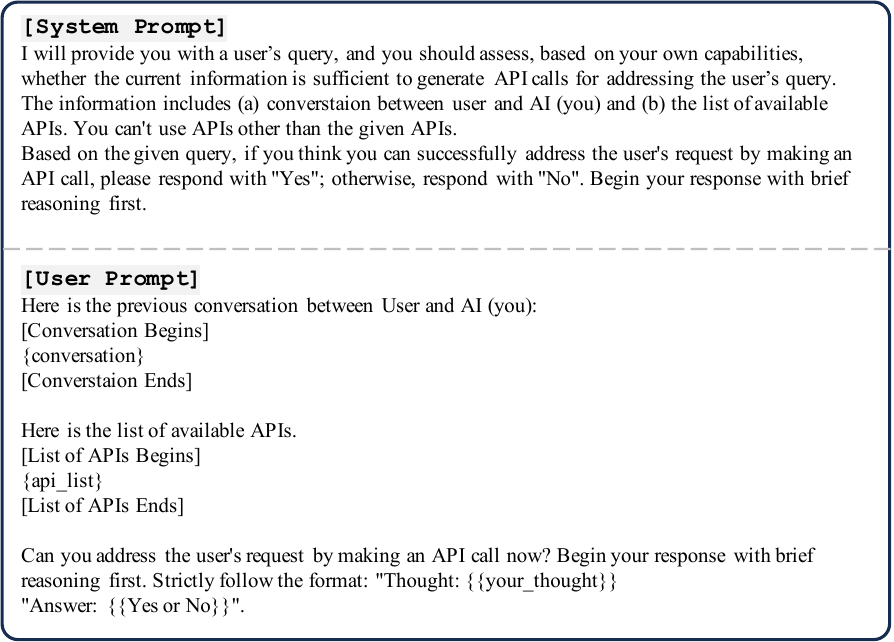}
  
\caption{CoT Experiments Prompt Template.}
\label{fig:app:cot_prompt_template}
 
\end{figure*}

\begin{figure*}
\centering
\includegraphics[width=\textwidth]{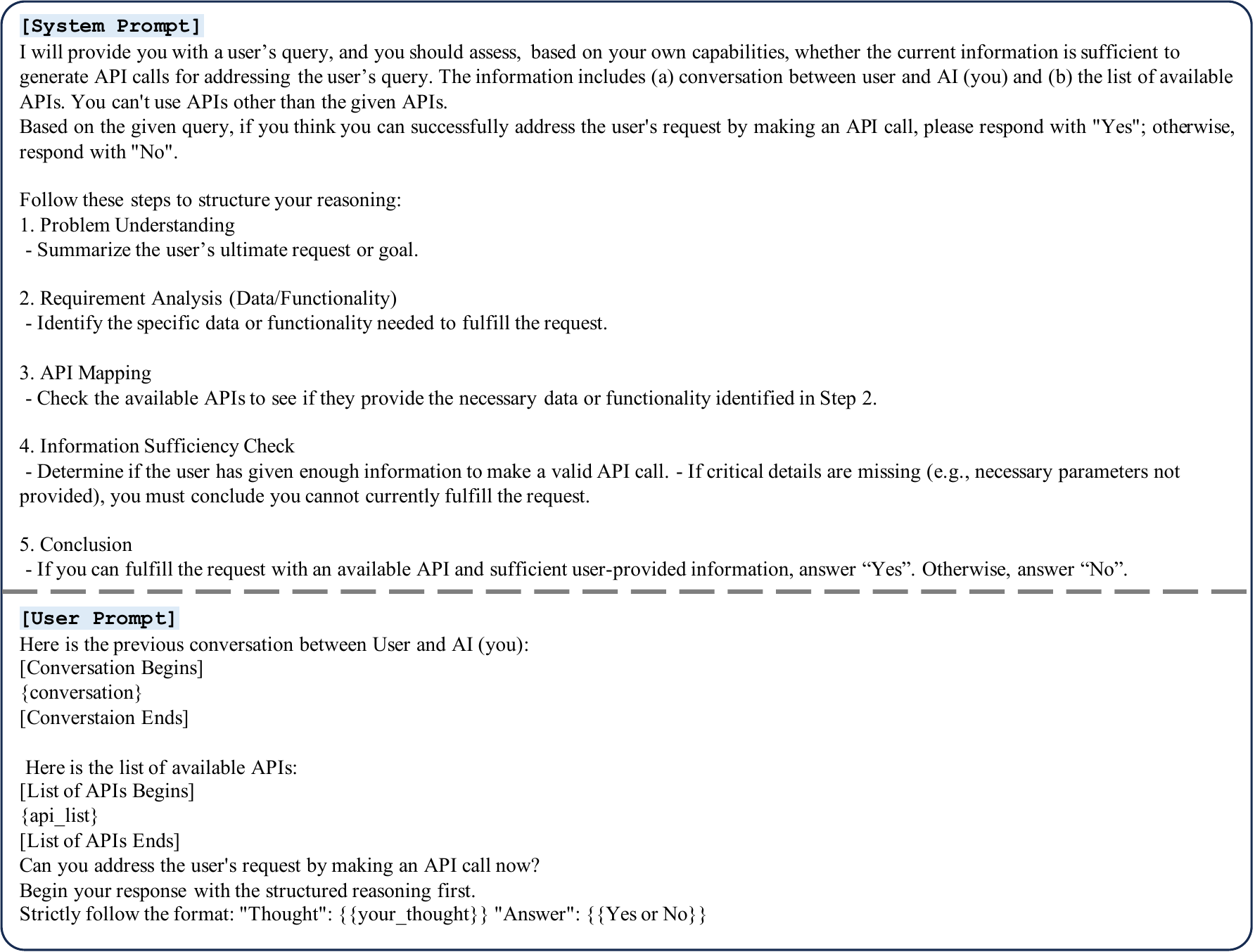}
  
\caption{Structured Verification Prompt Template.}
\label{fig:app:ours_prompt_template}
 
\end{figure*}

\begin{figure*}
\centering
\includegraphics[width=0.8\textwidth]{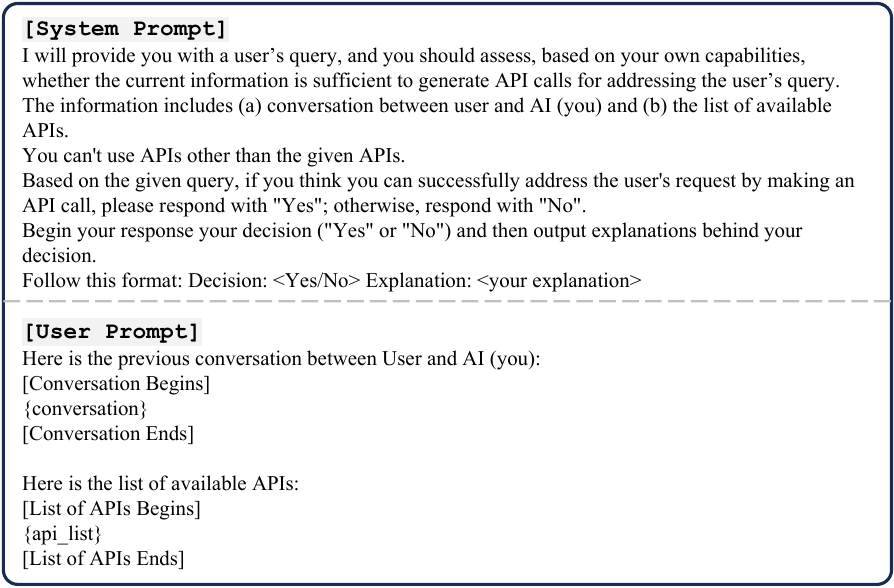}
  
\caption{Prompt template of explanation.}
\label{fig:app:task2_explanation_prompt_template}
 
\end{figure*}

\begin{figure*}
\centering
\includegraphics[width=0.8\textwidth]{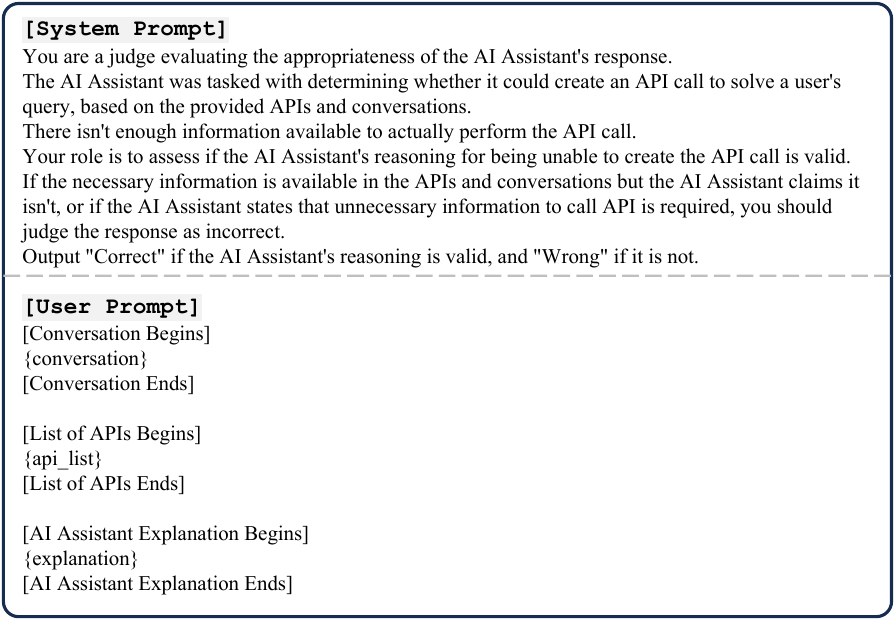}
  
\caption{Prompt template of Judge LLM.}
\label{fig:app:task2_judge_llm_prompt_template}
 
\end{figure*}

\begin{figure*}
\centering
\includegraphics[width=0.8\textwidth]{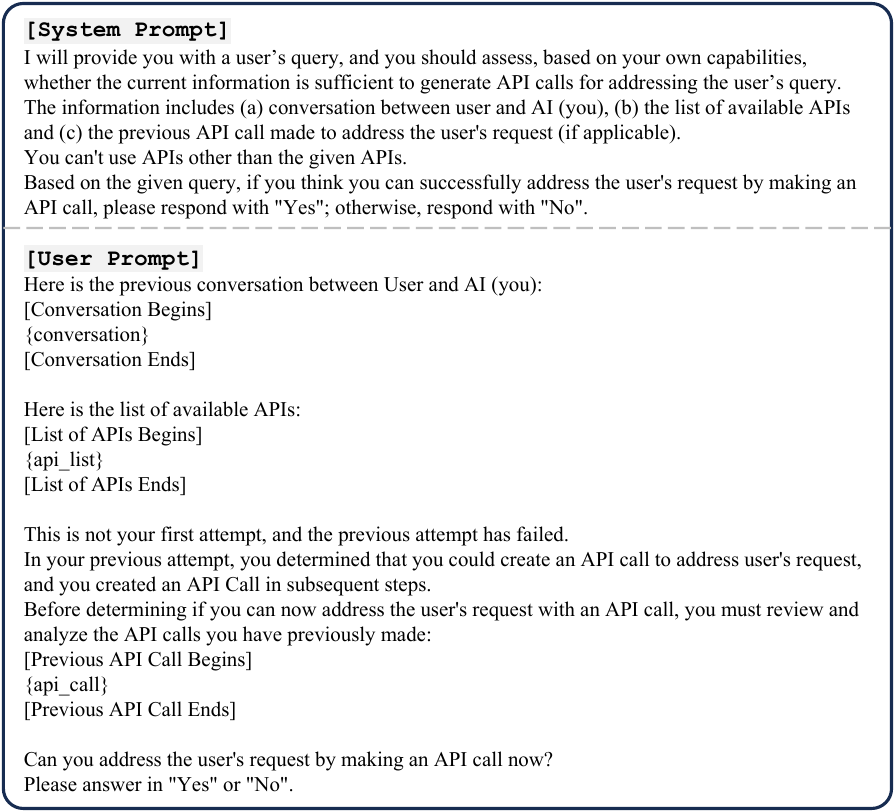}
\caption{Prompt Template of tool invocation feedback for API invocation error. We use the same prompt for error in utterance removal and the complete scenario. Both receive information about an erroneous API call result and are asked whether the API call is currently feasible.}
\label{fig:app:task3_error_prompt_template}
 
\end{figure*}

\begin{figure*}
\centering
\includegraphics[width=0.8\textwidth]{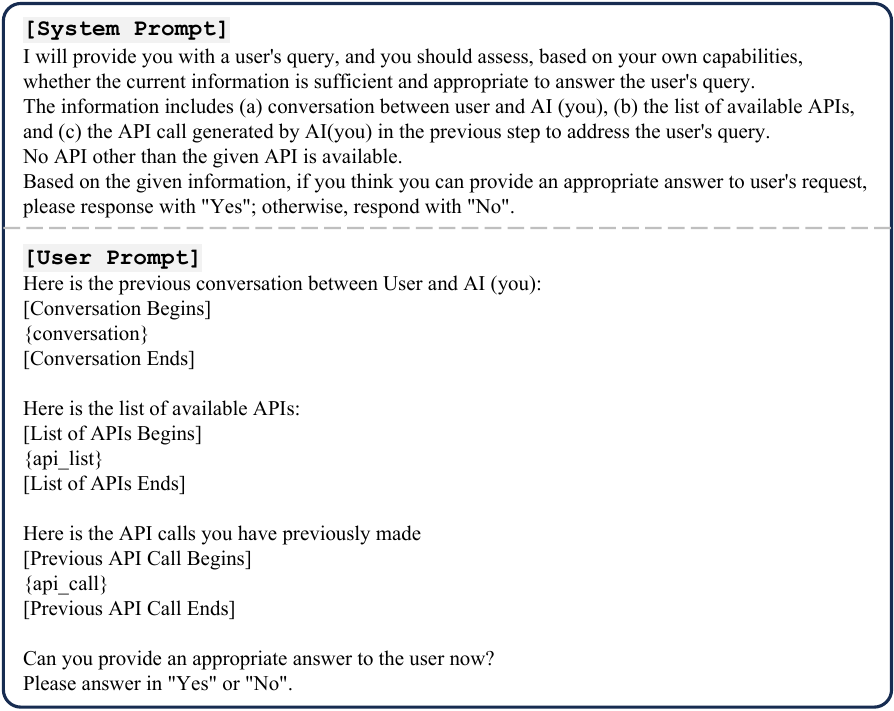}
\caption{Prompt Template of tool invocation feedback for hallucination.}
\label{fig:app:task3_hallucination_prompt_template}
\end{figure*}

\begin{figure*}
\centering
\includegraphics[width=0.8\textwidth]{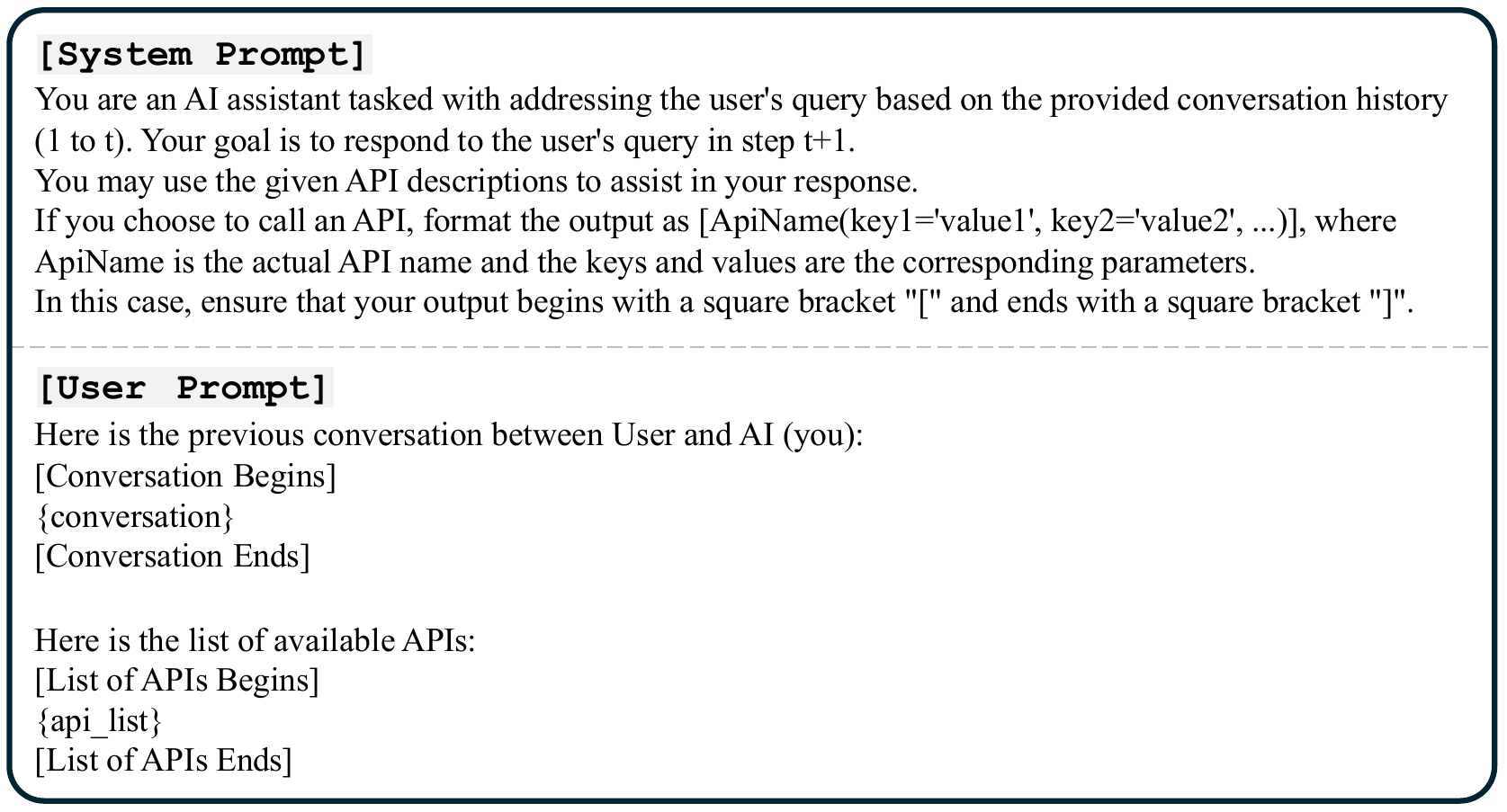}
\caption{Prompt Template of implicit evaluation of LLMs with free-form generation.}
\label{fig:app:free-form}
\end{figure*}

The text prompt used in the dataset construction is presented in 
Fig.~\ref{fig:app:data_construction_prompt_template}.
The text prompt used in the zero-shot experiments 
is presented in Fig.~\ref{fig:app:task1_prompt_template}.
The text prompts used in CoT and Structured Verification experiments are in 
Fig. \ref{fig:app:cot_prompt_template} and Fig.\ref{fig:app:ours_prompt_template}.
The text prompts used in the experiments of LLM explanation and LLM judgment are presented in Fig.~\ref{fig:app:task2_explanation_prompt_template} and Fig.~\ref{fig:app:task2_judge_llm_prompt_template}, respectively.
The text prompts used in self-verification are in Fig.~\ref{fig:app:task3_error_prompt_template} and Fig.~\ref{fig:app:task3_hallucination_prompt_template}. The text prompt used in implicit evaluation is in Fig.~\ref{fig:app:free-form}. 
When implementing few-shot prompting, we follow the approach of setting up interactions between a human and an AI assistant to provide examples~\cite{few-shot-prompts}.

% APIBank
% API
% Success

% Reproducibility Checklist

% ToolBench
% API
% Success

% APIBank
% API
% Fail

% ToolBench
% API
% Fail
 
% APIBank
% Utterance
% Success

% ToolBench
% Utterance
% Success

% APIBank
% Utterance
% Fail

% ToolBench
% Utterance
% Fail

% API Correct
 
% API Wrong
 
% User Correct
 
% User Wrong
 
% Data Construction
 
% Task1
 
% CoT
 
% Verification
 
% Task2
 
% Task3
 
\end{document}